# Physics-informed neural networks by Gradient-Guided Gaussian Adaptive Sampling (3GAS-PINNs)

Yousen Wang, [a] Wei Zhao [a,*]

[a] State Key Laboratory of Photon-Technology in Western China Energy, International Collaborative Center on Photoelectric Technology and Nano Functional Materials, Key Laboratory of Optoelectronic Technology of Shaanxi Province, Institute of Photonics & Photon Technology, Northwest University, Xi'an 710127, China

* Correspondence: zwbayern@nwu.edu.cn

**Abstract**

Physics-informed neural networks (PINNs) provide a mesh-free framework for solving partial differential equations, yet their performance in nonlinear problems is often limited by slow convergence, gradient imbalance, and insufficient resolution to capture localized intermittent structures such as shock waves[1]. These issues arise primarily from the use of fixed weights of loss and uniform collocation point distributions, which cannot adapt to the evolving complexity of the solution field during training.

To address these challenges, Gradient-Guided Gaussian Adaptive Sampling Physics-Informed Neural Networks (3GAS-PINNs) is proposed in this paper, which combines uniform probability distribution and Gaussian-smoothed probability distribution derived from the spatial gradients of solution, to maintain global constraint satisfaction as well as concentrating collocation points in regions of high gradient. Thus, intermittency structures like shock wave and solitons can be accurately captured.

The method is evaluated on three benchmark nonlinear problems, including one-dimensional forced Burgers equation, Korteweg-de Vries (KdV) equation and nonlinear Schrödinger equation, all of which exhibit steep gradients or strong nonlinearity. In comparison with baseline PINNs, 3GAS-PINNs can effectively promote the physical consistency in intermittent regions. The accuracy of the numerical simulation can be improved by a factor of up to 14.

## 1. Introduction

Nonlinear systems represent the most formidable yet pivotal research subjects in modern science and engineering. Under identical initial and boundary conditions, these systems exhibit a high degree of complexity and unpredictability. Mathematically and physically, this complexity manifests as the non-uniqueness of solutions—where multiple stable or metastable equilibrium states may coexist—and transitions in system behavior (such as the laminar-to-turbulent transition) as control parameters beyond critical thresholds. Furthermore, significant hysteresis effects imply that the system state is not solely determined by current inputs but is also path-dependent[2]. More crucially, nonlinear systems often feature pronounced multi-scale characteristics and parameter sensitivities; microscopic physical processes, such as molecular diffusion, can be amplified through nonlinear mechanisms to dictate macroscopic dynamics like shock wave formation or crack propagation[3]. Consequently, to deeply analyze these systems and elucidate their underlying macro-micro mechanisms, the development of efficient and high-precision numerical simulation methods is imperative.

Although grid-based traditional numerical methods, e.g. Finite Element Method (FEM)[4], Finite

Difference Method (FDM) [5], Finite Volume Method (FVM)[6], Direct Numerical Simulation (DNS)[7], Large-Eddy Simulation (LES)[8], Spectral Methods[9] and etc) perform well for linear, low-dimensional, and regular problems[10, 11], but capturing high-gradient structure requires excessive fine mesh in nonlinear system stimulation, which lead to exponentially increasing in computational cost, especially in high-Reynolds-number turbulence simulations[12].

To address the limitations of traditional grid-based methods, pure data-driven modeling based on Deep Neural Networks (DNNs)[13] has emerged prominently. These methods rely on statistical learning to extract latent correlations and physical mappings from massive, unstructured datasets. Nevertheless, data-driven approaches suffer from two critical defects: (1) An dependency on massive and high-quality labeled data, and (2) the inherent "black-box" nature of deep learning, which lacks interpretability and fails to explicitly incorporate physical laws or conservation principles[14].

To overcome the defects, the "data-driven + physics-driven" hybrid modeling paradigm has emerged, providing a mesh-free pathway. The most representative development in this field is the Physics-Informed Neural Networks (PINNs)[15]. The core philosophy of PINNs involves integrating governing PDEs into the loss function as regularization terms[15]. By leveraging Automatic Differentiation (AD)[16], PINNs precisely compute the derivatives of the network output with respect to input coordinates, thereby constructing residuals between the predicted solution and the physical model. Combined with data constraints from Boundary Conditions (BCs) and Initial Conditions (ICs), PINNs can approximate PDE solutions in a mesh-free manner. Despite their potential, standard PINNs focus primarily on single-physics problems and lack robust mechanisms for handling strong nonlinearities across multiple fields. Their original framework exhibits significant limitations in nonlinear Systems scenarios. On one hand, the loss function typically comprises fixed weighting of PDE residuals, BCs, and ICs. Fixed weights often lead to training conflicts between data fitting and physical constraints. For instance, the network may overfit boundary data to minimize boundary loss while neglecting internal dynamics, resulting in poor physical consistency[17, 18]. On the other hand, nonlinear problems involve intermittent regions where physical quantities vary drastically. Standard PINNs employ uniform or random sampling, which fails to allocate computational resources dynamically based on physical complexity. This imbalance results in redundant sampling in smooth regions and sparse sampling in critical areas, leading to the "smearing" of shock fronts and compromised accuracy.

In nonlinear physics simulations, uniform sampling typically leads to gradient lag and diminished accuracy in intermittent regions. Consequently, the optimization of sampling strategies is regarded as a critical breakthrough. The FAMAW-PINNs proposed by Wang et al. [4, 19]adopt an adaptive point-moving mechanism inspired by the firefly algorithm to dynamically adjust collocation point distribution. Branchini et al. introduced self-normalizing sampling[20], which reconstruct sampling probability density functions (PDFs) based on residual magnitude to prioritize high-residual locations. Gao et al. combined residuals with error indicators from finite element methods to generate PDFs for training set optimization[21]. Jiao et al. utilized a novel PDF blending regional residuals with Gaussian weights to achieve adaptive sampling[1, 22]. While these methods have effectively optimized collocation resource allocation and training convergence efficiency, they merely improve the sampling strategy and loss function of PINNs by minimizing the deviation between predicted and the reference solutions. In PINNs loss function, the gradient magnitudes of different terms (e.g., governing equations and boundary conditions) can differ drastically, causing the model to neglect intermittent regions (such as shockwaves) during training[23]. As a result, the neural network tends to learn low-frequency smooth solution first, making it difficult for the model to capture high-frequency physical features and thus losing physical consistency[24, 25].

Upon this context, the present study proposes Gradient-Guided Gaussian Adaptive Sampling Physics-Informed Neural Networks (3GAS-PINNs). By constructing a gradient-guided Gaussian probability density adaptive sampling strategy, this framework aims to enhance convergence efficiency and precision in nonlinear physics simulations. Specifically, it strengthens physical consistency in intermittent regions characterized by high gradients and strong nonlinearities. This paper details the theoretical foundation and algorithmic design of the proposed method, provides validation results across the Burgers equation[26], Korteweg-de Vries equation (KdV) equation[27], and Nonlinear Schrödinger equation[28], and offers an in-depth analysis of its advantages regarding improve physical consistency[29].

## 2. Gradient-Guided Gaussian Adaptive Sampling PINNs (3GAS-PINNs)

### 2.1 Physics-Informed Neural Networks

PINNs approximate the solutions of partial differential equations (PDEs) by transforming them into a continuous optimization problem. A deep neural network $u(\boldsymbol{x}, t; \theta)$ is employed as a surrogate model, integrating physical laws into the loss function to constrain the output[30]. Consider a general nonlinear PDE defined as

$$\mathcal{P}[u] = f(\boldsymbol{x}, t), (\boldsymbol{x}, t) \in \Omega \times [0, T] \tag{1}$$

where $u$ is the exact solution, $\mathcal{P}$ represents a nonlinear operator, $f(x, t)$ is the forcing function, $\boldsymbol{x}$ and $t$ denote the spatial and temporal coordinates, and $\Omega \times [0, T]$ is the spatio-temporal domain, where the dimension of $\Omega$ depends on the problem under investigation.

PINNs utilize Automatic Differentiation (AD) to compute the partial derivatives of the network output with respect to the input coordinates. This mechanism provides exact derivative values at any coordinate, avoiding the truncation errors inherent in grid-based numerical methods[16, 31].

The physics residual is defined as

$$R_\theta(\boldsymbol{x}, t) = \mathcal{P}[u(\boldsymbol{x}, t; \theta)] - f(\boldsymbol{x}, t) \tag{2}$$

where $u(\boldsymbol{x}, t; \theta)$ is the neural network approximation, $\theta$ represents the set of trainable parameters (e.g. weights and biases), and $R_\theta(\boldsymbol{x}, t)$ is the residual quantifying the discrepancy between the network output and the governing physical law or the reference solution.

The training of a PINNs is formulated as the minimization of a composite loss function $\mathcal{L}(\theta)$, which incorporates multi-objective regularization terms to ensure physical consistency across the entire spatiotemporal domain[24, 32]. This objective function is typically defined as a weighted sum of the physical constraint violation and the initial/boundary constraint violation[15]

$$L(\theta) = L_{res} + \ L_{ic} \ + \ L_{bc} \tag{3}$$

where $L_{res}$, $L_{ic}$, $L_{bc}$, are the loss components corresponding to the PDE residual, initial conditions, and boundary conditions, respectively. These loss components are quantified using the Mean Squared Error (MSE) over discrete sets of points, as

$$L_{res} = \frac{1}{N_r} \sum_{i=1}^{N_r} \left\| R_\theta(\boldsymbol{x}_r^i, t_r^i) \right\|^2 \tag{4}$$

where $N_r$ is the number of collocation points, and $(\boldsymbol{x}_r^i, t_r^i)$ represents the $i$-th collocation point within the computational domain $\Omega \times [0, T]$,with $x_r^i \in \Omega$ and $t_r^i \in [0, T]$.

The initial and boundary constraints are defined as follows[15]

$$L_{ic} = \frac{1}{N_{ic}} \sum_{j=1}^{N_{ic}} \left\| u(\boldsymbol{x}_{ic}^{j}, 0) - g_{ic}^{j} \right\|^2 \tag{5}$$

where $N_{ic}$ is the number of initial condition points, and $(\boldsymbol{x}_{ic}^{j}, t_{ic}^{j})$ denotes the $j$-th initial condition point, with $\boldsymbol{x}_{ic}^{j} \in \Omega$ and $t_{ic}^{j} = 0$

$$L_{bc} = \frac{1}{N_{bc}} \sum_{l=1}^{N_{bc}} \left\| \mathcal{B}[u](\boldsymbol{x}_{bc}^{l}, t_{bc}^{l}) - g_{bc}^{l} \right\|^2 \tag{6}$$

where $N_{bc}$ is the number of boundary condition points. and $(\boldsymbol{x}_{bc}^{l}, t_{bc}^{l})$ denotes the $l$-th boundary condition point, with $x_{bc}^{l} \in \partial\Omega$ and $t_{bc}^{l} \in [0, T]$.

In the standard PINNs, the weight coefficients are usually fixed. The optimization is performed using gradient-based algorithm (e.g. Adam optimizer in this study), to find the optimal parameters $\theta$ that satisfy the governing equations and constraints simultaneously. This mesh-free approach is particularly effective for systems with complex geometries where traditional mesh generation is unachievable.

To better capture localized intermittent structures, we propose an 3GAS-PINNs. It adjusts the standard objective function (Eq. 3) by dynamically tuning the loss landscape and collocation point distribution[33].

## 2.2 3GAS-PINNs algorithm

### *2.2.1 Deficiency of uniformly random sampling in PINNs*

Uniformly random sampling in PINNs usually fails to resolve fine structures with high gradients. Here, we take Burgers equation as an example to illustrate the potential issues of uniform sampling in PINNs. The governing equation is defined as follows[15]

$$\begin{cases} u_t + uu_x - \nu u_{xx} = 0, (x, t) \in [0,1] \times [0, T] \\ u(x, 0) = u_0(x), x \in [0,1] (\text{Initial Condition, IC}) \\ u(0, t) = u(1, t) = 0, t \in [0, T] (\text{Boundary} \cdot \text{Conditions,} \cdot \text{BCs}) \end{cases} \tag{7}$$

Where $u(x, t)$ denotes the localized fluid velocity as a function of the spatial coordinate $x$ and temporal coordinate $t$ within the bounded spatio-temporal domain $(x, t) \in [0,1] \times [0, T]$. For short, $u_t = \partial u / \partial t$ $u_x = \partial u / \partial x$ and $u_{xx} = \partial^2 u / \partial x^2$, respectively. This problem has an initial condition as $u_0(x) = \sin(\pi x)$ and Dirichlet boundary conditions. The kinematic viscosity coefficient is set to $\nu = 0.01/\pi$ to ensure the formation of shock wave[34]. In this section, we uniformly sample 5,000 random training points within the computational domain.

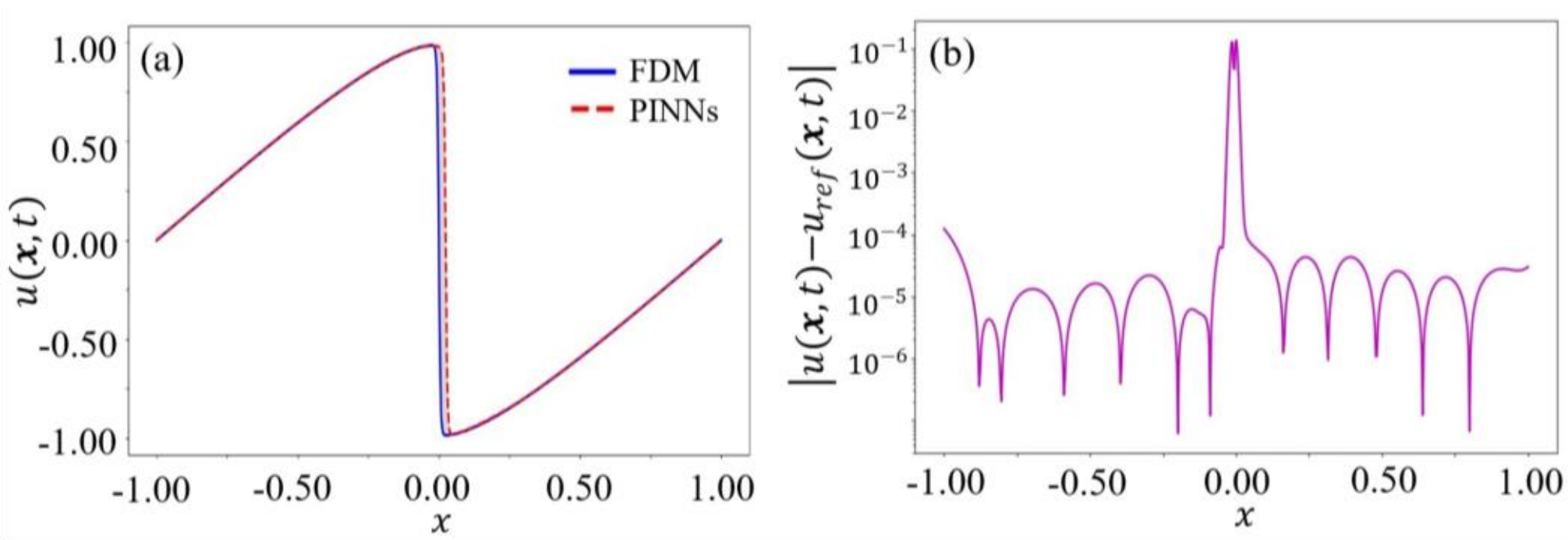

**Fig. 1** (a) Comparison between the predicted solution from PINNs and the FDM reference solution for Eq. (7) at $t = 0.5$ is presented. (b) $e_u$ defined in Eq. (14) between the predicted solution by PINNs and the reference solution by FDM for Eq. (7) at $t = 0.5$ is presented.

From Fig. 1, it can be seen the predicted solution by PINNs exhibits a significant deviation from the reference solution. This deviation arises from the formation of intermittent regions (e.g., shock wave fronts) in the spatiotemporal domain, induced by nonlinear advection in the governing equations. Uniformly random sampling points fail to capture the influence of localized intermittent regions on the training process adequately. Intermittent regions with steep gradients demand high-fidelity simulation, yet the uniformly random sampling strategy allocates an insufficient number of sampling points in these regions to enable high-precision simulation. This deficiency directly gives rise to substantial predictive errors in PINNs[35].

### *2.2.2 Gradient-Guided Gaussian Adaptive Sampling*

To improve the simulation accuracy of localized intermittent structures (e.g., shock waves, rogue waves and etc), the sampling strategy must dynamically reallocate computational resources. For this purpose, we develop a Gradient-Guided Gaussian Adaptive Sampling method depending on the spatial variations of $u_{\theta_k}(\boldsymbol{x})$. Here, $\theta_k$ represents the neural network parameters at the $p$-th resampling interval. The method is realized through a three-step mathematical framework, as illustrated in the schematic in Fig. 2.

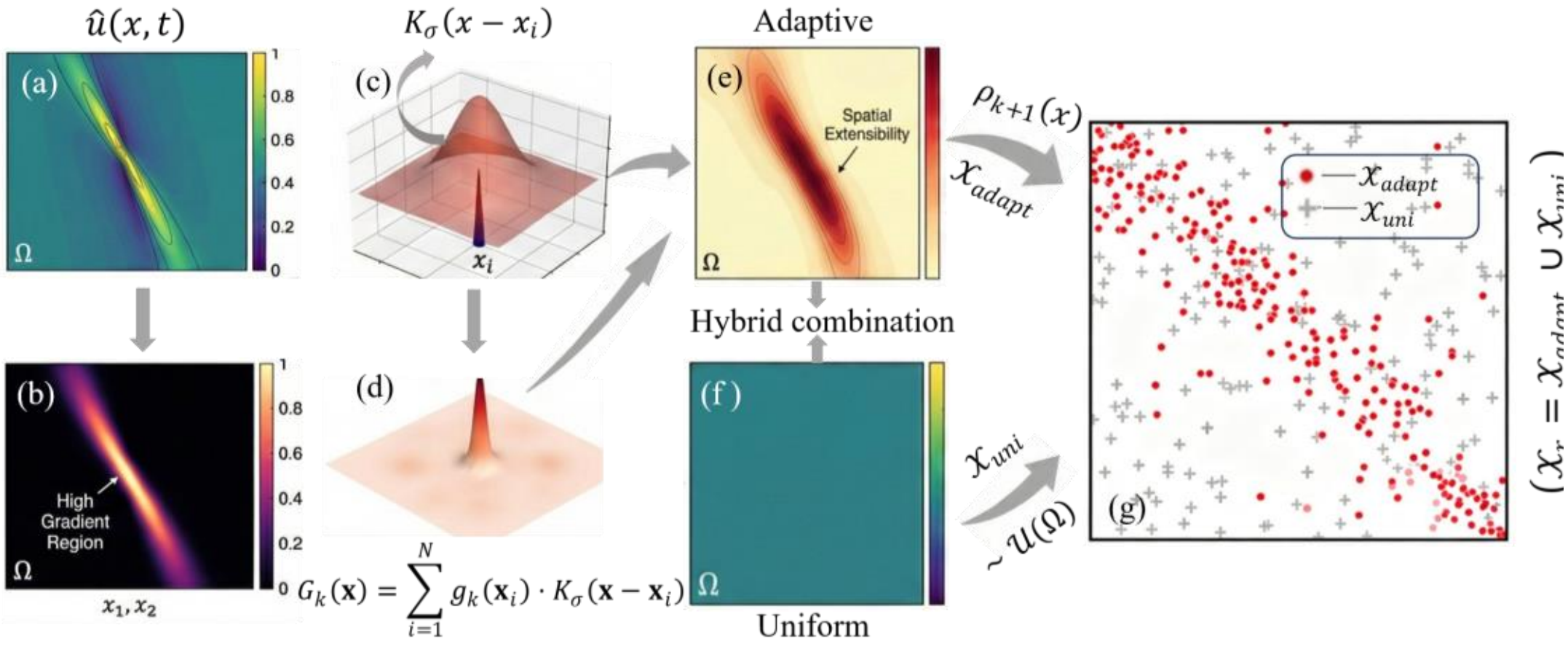

**Fig. 2** Schematic of the Gradient-Guided Gaussian Adaptive Sampling

**Step 1: Evaluation of the magnitude of gradient field (Fig. 2(a), (b))**

At the $k$-th step, the magnitude of gradient field of $u_{\theta_k}(\boldsymbol{x})$, defined as $g_k(\boldsymbol{x})$, is calculated at a discrete set of reference points $\{x_i\}_{i=1}^{N} \subset \Omega$. Based on the current step predicted solution **(Fig. 2(a)**), the local physical intensity for a $d$-dimensional spatial domain is quantified as[36]:

$$g_k(x_i) = \sqrt{\sum_{j=1}^{d} \left(\frac{\partial u_{\theta_k}}{\partial x_i^j}\right)^2} \tag{8}$$

where $\partial u_{\theta_k} / \partial x_i^j$ is evaluated via automatic differentiation. This field directly reflects the spatial rate of change of the physical quantity, marking high Gradient regions **(Fig. 2 (b))** of potential rapid transition.

**Step 2: Construction of the Gaussian Probability Density Field ((Fig. 2(c)-(e))**

Direct sampling according to $g_k(x_i)$ tends to cluster sampling points at the extremum points. This disrupts the spatial continuity of the sampling distribution, thereby compromising optimization convergence and inducing numerical instability. To ensure spatial extensibility **(Fig. 2(e))** and scalability, we construct a continuous smoothed intensity field $G_k(\boldsymbol{x})$ by convolving discrete gradients with a Gaussian kernel[37] $K_\sigma$ **(Fig. 2(c))**, which is given by:

$$G_k(\boldsymbol{x}) = \sum_{i=1}^{N} g_k(\boldsymbol{x_i}) \cdot K_\sigma(\boldsymbol{x} - \boldsymbol{x_i}) \tag{9}$$

The Gaussian kernel $K_\sigma$ is defined as:

$$K_\sigma(\boldsymbol{x} - \boldsymbol{x_i}) = \frac{1}{(2\pi\sigma^2)^{d/2}} \exp\left(-\frac{\| \boldsymbol{x} - \boldsymbol{x_i} \|^2}{2\sigma^2}\right) \tag{10}$$

The result is a smoothed manifold representing the importance of each region **(Fig. 2(d))**, Whether a region needs refined sampling depends on both the absolute gradient magnitude and the relative significance of such gradient variations against the global field background. The selection of $\sigma$ is crucial. An over-wide kernel causes overly uniformed sampling, whereas an over-narrow one leads to excessive clustering of collocation points, resulting in isolated cluster distributions.

To achieve adaptive sampling, $\sigma$ is determined as following. Let the mean gradient and mean solution be defined as $\bar{g}_k = \frac{1}{N}\sum_{i=1}^{N} g_k(\boldsymbol{x_i})$ and $\bar{\hat{u}}_k = \frac{1}{N}\sum_{i=1}^{N} \hat{u}_{\theta_k}(\boldsymbol{x_i})$, respectively. The standard deviation of the gradient field is $g_{k,std} = \sqrt{\frac{1}{N}\sum_{i=1}^{N} (g_k(\boldsymbol{x_i}) - \bar{g}_k)^2}$, and the standard deviation of the solution is $u_{k,std} = \sqrt{\frac{1}{N}\sum_{i=1}^{N} (\hat{u}_{\theta_k}(\boldsymbol{x_i}) - \bar{\hat{u}}_k)^2}$ . By analogy with the definition of Taylor microscale which represents characteristic eddy size in turbulence, the adaptive kernel width is defined as:

$$\sigma = A \cdot \frac{u_{k,std}}{g_{k,std}} \tag{11}$$

where $A$ is an empirical constant to be determined. This formulation ensures that the influence of high-density points extend to their neighbors, creating a continuous sampling manifold.

**Step 3: Normalization and Hybrid Sampling (Fig. 2(e)-(g))**

The new probability density function ($\rho_k(x)$) is obtained by normalizing the smoothed intensity field over the spatial domain $\Omega$:

$$\rho_{k+1}(x) = \frac{G_k(x)}{\int_\Omega G_k(x')dx'} \tag{12}$$

Meanwhile, it is subject to $\rho_{k+1}(x) \geq 0$ and $\int_\Omega \rho_{k+1}(x)dx = 1$. To avoid insufficient sampling points in smooth regions, a hybrid sampling strategy is employed. Given the total number of collocation points $N$ and adaptive portion $\eta \in [0,1]$, the training set $\mathcal{X}_r$ **(Fig. 2(g))** for the next iteration is defined as:

$$\mathcal{X}_r = \mathcal{X}_{adapt} \cup \mathcal{X}_{uni} \tag{13}$$

where the adaptive subset $\mathcal{X}_{adapt}$ which has $\lfloor \eta N \rfloor$ points [14] is sampled according to $\rho_{k+1}(x)$ **(Fig. 2(e))**, while the uniform subset $\mathcal{X}_{uni}$ which has $N - \lfloor \eta N \rfloor$ points is sampled uniformly **(Fig. 2(f))**. This ensures intermittent regions are resolved with high density, while maintaining the PDE's global constraints across the entire domain. And the gradient-guided Gaussian Adaptive Sampling Algorithm is presented in Table 2.

**Table 1.** Gradient-Guided Gaussian Adaptive Sampling Algorithm

| **Algorithm2: Gradient-Guided Gaussian Adaptive Sampling** |
|---|
| **Input:** Model $\hat{u}_{\theta_k}$, Adaptive ratio $\eta$, frequency $K$, scaling $A$, total points $N$, *warmup* |
| **Output:** Optimized collocation set $\mathcal{X}_r$. |
| 1. **if** $epoch < warmup$ **then** |
| 2. Sample $\mathcal{X}_{uni} \sim \mathcal{U}(\Omega \times [0, T])$ uniformly. |
| **3. else if** *epoch* (mod $K$) = = 0 **then** |
| 4. Calculate local gradients $g_k(\mathrm{x}_i)$ at reference points. |
| 5. Update adaptive Gaussian kernel width $\sigma$. |
| 6. Construct the smoothed intensity field $G_k(x)$. |
| 7. Normalize $G_k(x)$ to define the probability density $\rho_{k+1}(x)$. |
| 8. **Resample adaptive points:** $\mathcal{X}_{adapt} \leftarrow$ sample $\lfloor \eta N \rfloor$ points from $\rho_{k+1}(x)$. |
| 9. **Resample uniform points:** $\mathcal{X}_{uni} \leftarrow$ sample $N - \lfloor \eta N \rfloor$ points from $\mathcal{U}(\Omega)$. |
| 10. **Update training set:** $\mathcal{X}_r = \mathcal{X}_{adapt} \cup \mathcal{X}_{uni}$. |
| **11. end if** |

The flowchart of the 3GAS-PINNs algorithm is shown in Fig.3. During training, the network takes spatial–temporal coordinates $(\boldsymbol{x}, t)$ as inputs and outputs the predicted solution $u(\boldsymbol{x}, t; \theta)$. The total loss $L(\theta)$ is composed of the PDE residual loss $L_{\mathrm{res}}$, initial condition loss $L_{\mathrm{ic}}$, and boundary condition loss $L_{\mathrm{bc}}$, the collocation points are adaptively resampled according to the Gradient-Guided Gaussian Adaptive Sampling Algorithm after a *warmup* stage, focusing more points on intermittent regions. The optimizer iteratively updates the network parameters $\theta$ using the dynamically weighted loss and the refined sampling set, until the model satisfies the physical constraints and achieves high precision of the spatiotemporal field.

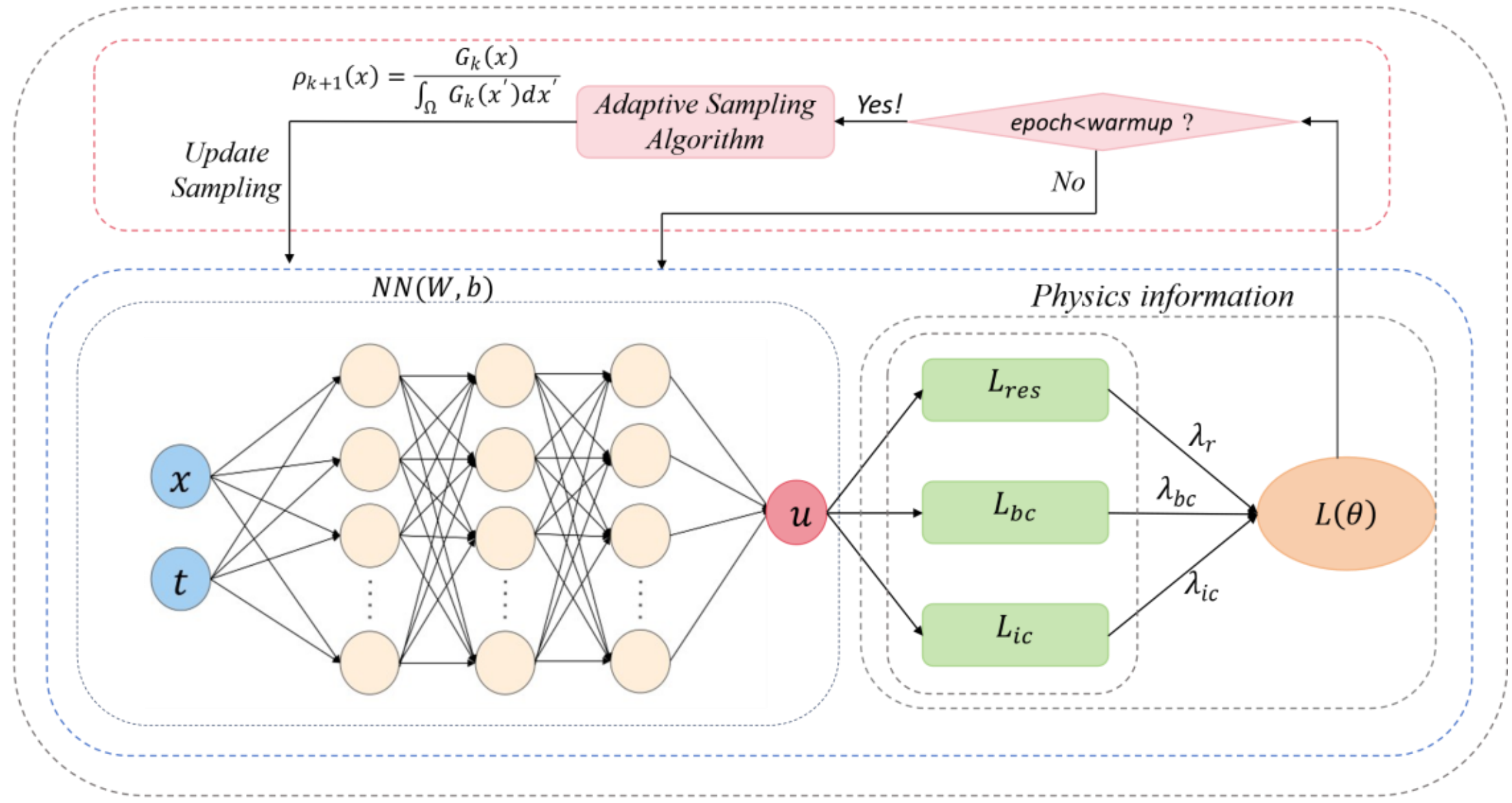

**Fig. 3** Flowchart of the 3GAS-PINNs algorithm

## 3. Numerical simulation results

To verify the performance of 3GAS-PINNs, we conduct tests on three benchmark problems: one-dimensional (1D) Burgers equation without/with forcing, Korteweg-de Vries equation, and nonlinear Schrödinger equation. This section elaborates on the evaluation metrics, experimental settings, and the corresponding results and analysis for the three benchmark tests in the numerical simulation experiments.

### 3.1 Evaluation metrics

To intuitively quantify the point-wise deviation between model predictions and reference solutions, we adopt two quantitative indicators including the absolute solution error and absolute gradient error as

(1) Absolute solution error

The absolute solution error is denoted as $e_u$, which measures the point-wise discrepancy between the predicted physical quantity and the reference solution. Its mathematical expression is:

$$e_u(x_i, t_s) = \sum_{j=1}^{d} |u(\boldsymbol{x_i}, t_s) - u_{ref}(\boldsymbol{x_i}, t_s)| \tag{14}$$

where $u(\boldsymbol{x}_i, t_s)$ and $u_{ref}(\boldsymbol{x}_i, t_s)$ denote the predicted solution and the reference solution at spatial point $\boldsymbol{x_i}$ and physical time instant $t_s$ respectively, with $(\boldsymbol{x}_i, t_s)$ referring to the discrete coordinates within the computational domain. Here, the index $s$ in $t_s$ refers to the discrete physical time levels of the PDE solution, describing the temporal evolution of the solution field in a single prediction process. It is independent of the training epochs of the neural network.

(2) Absolute gradient error

The absolute gradient error is denoted as $e_{\nabla u}$, which characterizes the point-wise deviation between the gradient of the predicted solution and that of the reference solution. Its mathematical expression is:

$$e_{\nabla u}(x_i, t_s) = \sum_{j=1}^{d} \left| \frac{\partial u(\boldsymbol{x_i}, t_s)}{\partial x^j} - \frac{\partial u_{ref}(\boldsymbol{x_i}, t_s)}{\partial x^j} \right| \tag{15}$$

where $\frac{\partial u(x_i,t_s)}{\partial x^j}$ and $\frac{\partial u_{ref}(x_i,t_s)}{\partial x^j}$ denote the spatial gradients of the predicted and reference solutions at the spatial point $\boldsymbol{x_i}$, time instant $t_s$, and spatial dimension $j$, respectively. Here, $d$ is the spatial dimension of the problem.

Furthermore, Relative $L_2$ error and Relative gradient $L_2$ error have been calculated to rigorously evaluate the accuracy and robustness of 3GAS-PINNs for resolving local intermittent structures, as

(3) Relative $L_2$ error

$$\varepsilon_{L_2} = \frac{\sqrt{\sum_{i=1}^{n_x} \sum_{s=1}^{n_t} e_u(x_i, t_s)^2}}{\sqrt{\sum_{i=1}^{n_x} \sum_{s=1}^{n_t} \left|u_{ref}(\boldsymbol{x}_i, t_s)\right|^2}} \tag{16}$$

where $n_x$ and $n_t$ are the number of spatial and temporal evaluation points over the computational domain, with $(\boldsymbol{x}_i, t_s)$ referring to the discrete coordinates within the computational domain.

(4) Relative gradient $L_2$ error

Capturing intermittent structures requires high-accuracy computation on the gradient of physical quantity. The overall deviation in gradients between the predicted and exact solutions reflects whether the predicted solutions capture the gradient information of the model accurately, which can be evaluated by a relative gradient $L_2$ error ($\varepsilon_{\mathrm{grad},L_2}$) as:

$$\varepsilon_{grad,L_2} = \frac{\sqrt{\sum_{i=1}^{n_x} \sum_{s=1}^{n_t} e_{\nabla u}(x_i, t_s)^2}}{\sqrt{\sum_{i=1}^{n_x} \sum_{s=1}^{n_t} \sum_{j=1}^{d} \left| \frac{\partial u_{ref}(\boldsymbol{x_i}, t_s)}{\partial x^j} \right|^2}} \tag{17}$$

These metrics are designed to quantify the relative discrepancies of the global field as well as the model's capability to resolve local intermittent structures.

### 3.2 Experimental settings

*3.2.1 Implementation Environment*

All simulations are implemented in Python 3.10 with the PyTorch library. Computations are performed on an AMD Ryzen 9 9950X CPU and 96 GB of DDR5 RAM running at 5600 MT/s, accelerated by an NVIDIA RTX 4090 GPU supporting CUDA 12.6, which accelerates the backpropagation process and the automatic differentiation required for compute the residuals.

*3.2.2 Neural Network Training configurations and Training Parameters*

The base neural network model is a fully connected deep neural network, which consists of 4 hidden layers with 64 neurons in each layer. The Tanh activation function is employed for all hidden layers, as its smoothness enables the stable computation of high-order derivatives during automatic differentiation.

Network parameters are optimized via the Adam algorithm with an initial learning rate of $10^{-3}$. A learning rate scheduler is applied to enhance convergence in the late training stage, halving the learning rate every 10,000 epochs. The model is trained for a total of 50,000 iterations, a duration sufficient to stabilize the weights and position the adaptive sampling points near the peaks of intermittent regions.

To facilitate the model in capturing the fundamental physical gradient characteristics of the solution, the first 2000 iterations are designated as the warm-up phase. Following the warm-up, mixed sampling is activated with the adaptive ratio $\eta$. The gradient-guided adaptive resampling process is implemented every 1000 iterations.

### 3.3 Burgers equation

To evaluate the capability of 3GAS-PINNs in solving nonlinear problems with discontinuous gradients and shock waves, we investigate the shock wave evolution governed by the Burgers equation in Eq. (7). A total of 5000 sampling points is employed for numerical simulations. The results are shown in Fig. 4. The reference shock wave solution of Eq. (7) is plotted in Fig. 4(a) according to the FDM benchmark. The shock wave structure locates at the temporal–spatial domain shows a drastic variation of $|u_x|$. To explicitly capture these structures, reserving their high-gradient feature, more sampling points are required in the intermittent regime. Fig. 4(b–f) show how $\eta$ controls the distribution of sampling points by the adaptive sampling. It can be clearly observed, when $\eta \geq 0.7$, the sampling points concentrate towards the intermittent regime (shock wave region). This is consistent to our expectation on 3GAS-PINNs, supporting the effectiveness.

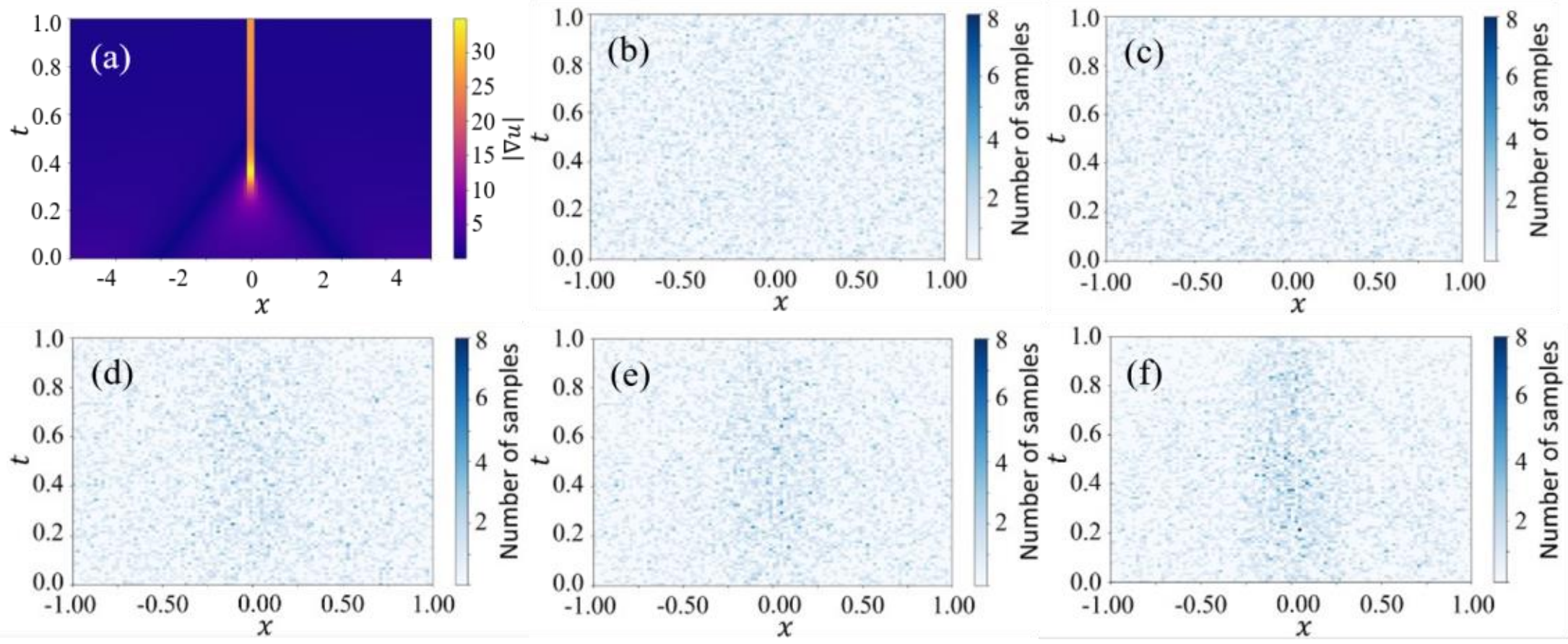

**Fig. 4** (a) Distribution of gradient magnitude of the reference solution by FDM at $t = 0.5$ of the Burgers equation (7). (b) to (f) correspond to the sampling point distributions of the final resampling for the numerical simulation by 3GAS-PINNs at $t = 0.5$, $A = 1$ of the Burgers equation (7), with the adaptive update ratio $\eta$ set to 0.1, 0.3, 0.5, 0.7, and 0.9, respectively.

The influence of $\eta$ on simulation accuracy has been summarized in Table 2. It can be seen, when $\eta = 0.7$, $\varepsilon_{L_2}$ and $\varepsilon_{grad,L_2}$ are $1.33 \times 10^{-3}$ and $7.13 \times 10^{-2}$ respectively. They are only 14.6% and 86.2% of those in the PINNs ($9.11 \times 10^{-3}$ and $8.27 \times 10^{-2}$). This is equivalent to a factor of 2~7 improvement on simulation accuracy. The results support that the 3GAS-PINNs is effective in solving shock wave problems governed by the Burgers equation.

**Table 2.** Results of 3GAS-PINNs for simulating Equation (7) when $A = 1$ with different values of $\eta$

| $\eta$ | $L(\theta)$ $(10^{-3})$ | $L(\theta)$ $(10^{-4})$ | $L(\theta)$ (Finally) | $\varepsilon_{L_2}$ | $\varepsilon_{grad,L_2}$ |
|---|---|---|---|---|---|
| 0.1 | 4851 | 10361 | 1.13 E-05 | 1.76 E-03 | 7.13 E-02 |
| 0.3 | 5751 | 13875 | 8.82 E-06 | 1.72 E-03 | 7.16 E-02 |
| 0.5 | 10094 | 21011 | 5.73 E-06 | 1.45 E-03 | 7.14 E-02 |
| 0.7 | 4900 | 16819 | 6.96 E-06 | **1.33 E-03** | **7.13 E-02** |
| 0.9 | 14484 | 45198 | 5.95 E-05 | 1.14 E-02 | 1.32 E-01 |
| PINNs | 15626 | - | 2.86 E-04 | 9.11 E-03 | 8.27 E-02 |

We further explore the influence of scaling constant $A$ of the adaptive density function in Eq. (11) on simulation accuracy. As shown in Fig. 5(a-f), sampling density functions are highly consistent to the distribution of $|u_x|$ in Fig. 4(a). When $A$ keeps increasing, $\rho$ becomes smeared. The intermittent structure becomes invisible when $A \geq 1$, The sampling point distribution also reflects the same tendency of variation. Intuitively, a small $A$, e.g. $A \leq 1$, should provide a better depiction on the intermittent structure, with higher simulation accuracy.

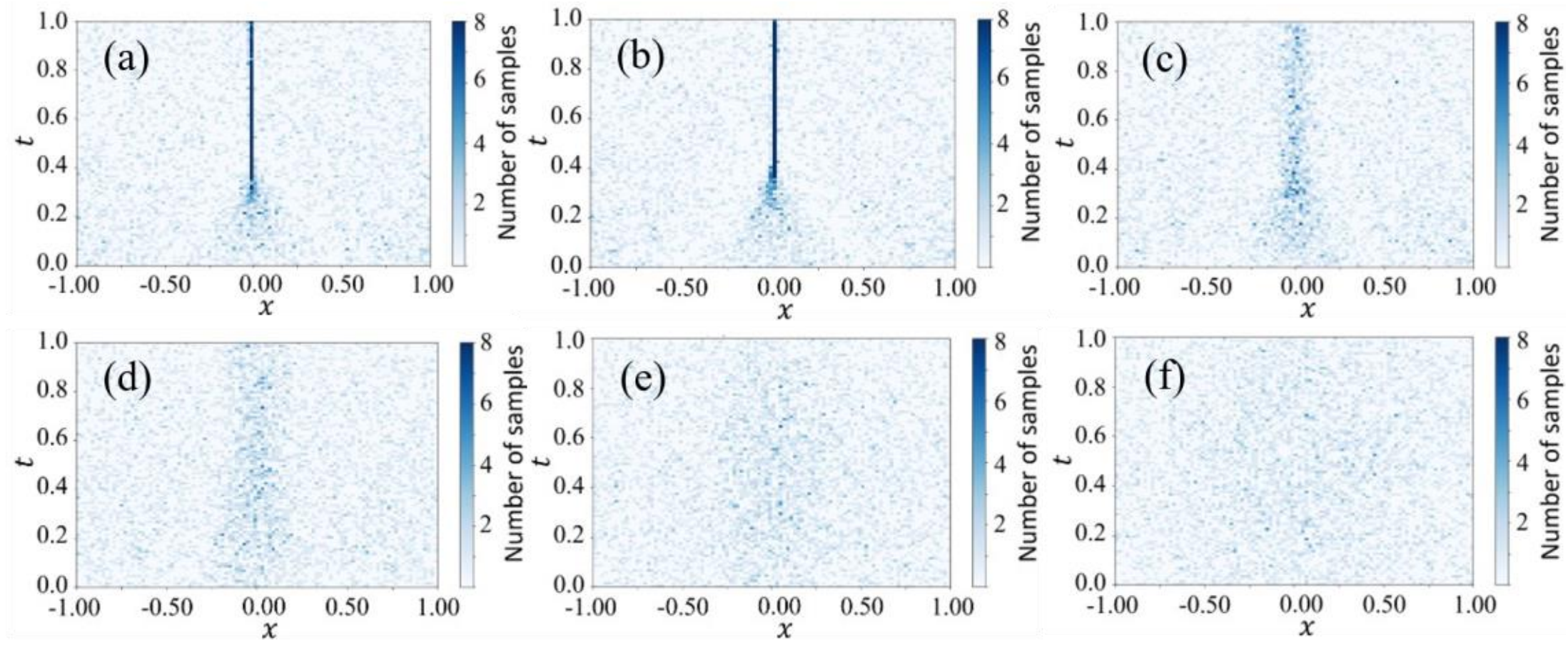

**Fig. 5** presents the final sampling point distribution map corresponding to the numerical simulation of the Burgers equation (7) by the 3GAS-PINNs at $t = 0.5$ with $\eta = 0.7$. (a)–(f) correspond to the final sampling point distributions, with the scaling constant A of σ set to 0.05, 0.1, 0.25, 0.5, 1, and 2, respectively.

However, the results are counterintuitive. Table 3 shows the optimal $A = 1$ provides the lowest $\varepsilon_{L_2}$ and $\varepsilon_{grad,L_2}$ respectively. It balances the simulation accuracy in both intermittent and smooth regimes. In contrast, when $A = 0.1$, sampling points are concentrated near the shock wave in the intermittent regime, causing a sparse distribution in the smooth regime and low accuracy up there. Relative to conventional PINNS, 3GAS-PINNs takes 4900 and 16819 epochs to reach $10^{-3}$ and $10^{-4}$ loss, respectively. The former one is only 31% of that by PINNs.

**Table 3.** Results of the 3GAS-PINNs for simulating Equation (7) when $\eta = 0.7$ with different values of $A$

| $A$ | $L(\theta)$ $(10^{-3})$ | $L(\theta)$ $(10^{-4})$ | $L(\theta)$ (Finally) | $\varepsilon_{L_2}$ | $\varepsilon_{grad,L_2}$ |
|---|---|---|---|---|---|
| 0.05 | 37115 | - | 4.52 E-04 | 1.27 E-01 | 1.04 E-00 |
| 0.10 | 24917 | - | 1.03 E-03 | 9.95 E-02 | 8.79 E-01 |
| 0.25 | 17917 | 24322 | 2.15 E-05 | 3.11 E-03 | 7.60 E-02 |
| 0.50 | 15976 | 34363 | 1.55 E-05 | 1.64 E-03 | 7.13 E-02 |
| 0.90 | 14527 | 32188 | 1.82 E-05 | 1.99 E-03 | 7.20 E-02 |
| 0.95 | 9614 | 19161 | 4.68 E-06 | 1.65 E-03 | 7.22 E-02 |
| 1.00 | 4900 | 16819 | 4.76 E-06 | **1.33 E-03** | **7.13 E-02** |
| 1.05 | 6403 | 28764 | 1.19 E-05 | 1.51 E-03 | 7.13 E-02 |
| 1.10 | 11484 | 19487 | 1.01 E-05 | 1.77 E-03 | 7,17 E-02 |
| 2.00 | 10065 | 22324 | 9.92 E-06 | 1.81 E-03 | 7.21 E-02 |
| PINNs | 15626 | - | 2.86 E-04 | 9.11 E-03 | 8.27 E-02 |

3GAS-PINNs is capable of generating a refined spatial distribution of sampling points based on the gradient profile. It is an advantage to capture the intermittent structures, however, may also lead to an insufficient number of sampling points elsewhere. Thus, the total number of sampling points is also crucial for simulation accuracy. Quantitative comparisons of $\varepsilon_{L_2}$ and $\varepsilon_{grad,L_2}$ in both 3GAS-PINNs and conventional PINNs have been made in Fig. 6. The results illustrate clearly that the 3GAS-PINNs achieves consistently lower $\varepsilon_{L_2}$ and $\varepsilon_{grad,L_2}$ than PINNs across all sampling budgets. In particular, both $\varepsilon_{L_2}$ and $\varepsilon_{grad,L_2}$ can be reduced by nearly one order of magnitude. Or, we can use only 200 sampling points to achieve comparable accuracy from PINNs which requires more than 2000 sampling points. This fully verifies that the 3GAS-PINNs can effectively improve the simulation accuracy of shock wave structures under limited computational resources.

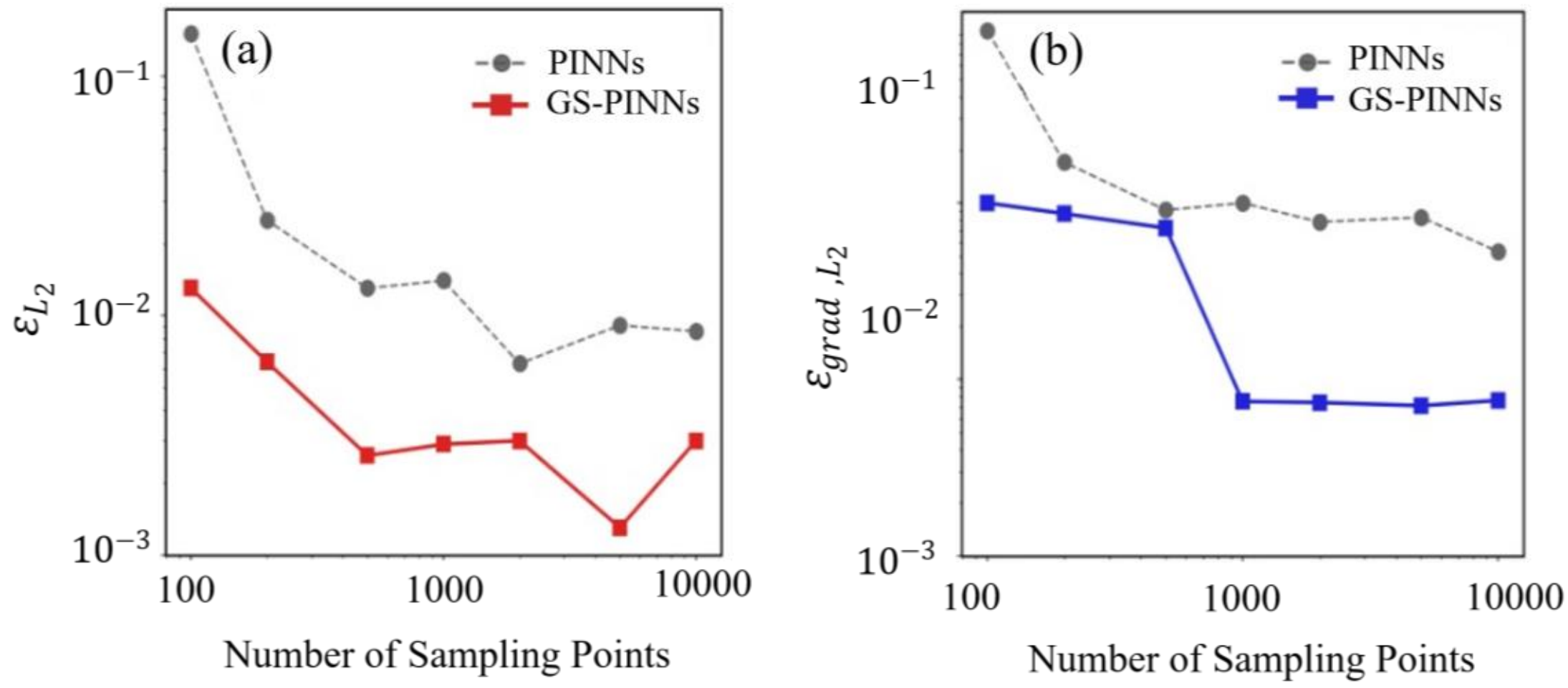

**Fig. 6** Influence of training point number on the errors of solving the Burgers equation (7) at $t = 0.5, A = 1$ and $\eta = 0.7$ by 3GAS-PINNs and PINNs. (a) $\varepsilon_{L_2}$. (b) $\varepsilon_{grad,L_2}$.

We systematically compare the predictive performance of conventional PINNs and the proposed 3GAS-PINNs with optimized hyperparameters ($\eta = 0.7, A = 1$) for solving the Burgers equation (7). Superficially, the predicted solutions by PINNs (Fig. 7 (a1)) and 3GAS-PINNs (Fig. 7 (a2)) are consistent to the reference solution by FDM. However, the loss functions demonstrate significant difference.

As plotted in Fig. 7 (b1) and (b2), the final training loss of PINNs converges to $2.81 \times 10^{-4}$, whereas 3GAS-PINNs reduces the terminal loss down to $4.76 \times 10^{-6}$, achieving a drop of nearly two orders of magnitude. In terms of spatial absolute error shown in Fig.7 (c1) and Fig.7 (c2), baseline PINNs produces prominent error patches surrounding the shock wave front, while 3GAS-PINNs effectively suppresses by adaptively enriching collocation points around high-gradient intermittent regions. Quantitatively, the $\varepsilon_{L_2}$ decreases from $9.11 \times 10^{-3}$ (PINNs) to $1.33 \times 10^{-3}$ (3GAS-PINNs), corresponding to an accuracy is improved by around 7 times. A consistent improvement can also be observed from the absolute gradient error distributions in Fig.7 (d1) and Fig.7 (d2). Quantitatively, the $\varepsilon_{grad,L_2}$ decreases from $8.27 \times 10^{-2}$ (PINNs) to $7.13 \times 10^{-2}$ (3GAS-PINNs), yielding a 16% accuracy improvement. These quantitative and visual evidences collectively verify that the 3GAS-PINNs can efficiently allocate limited computational resources toward intermittent structures and substantially boost the overall numerical precision under identical collocation point budgets.

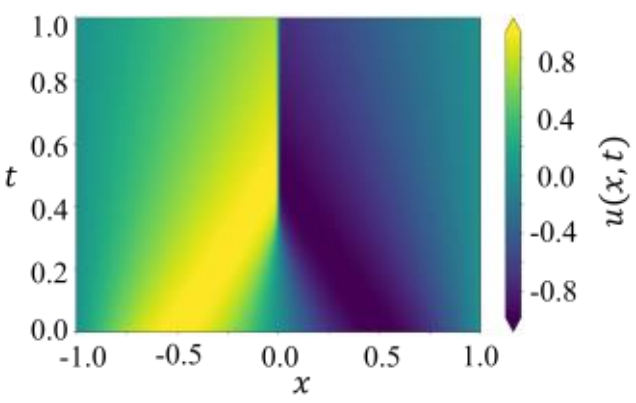

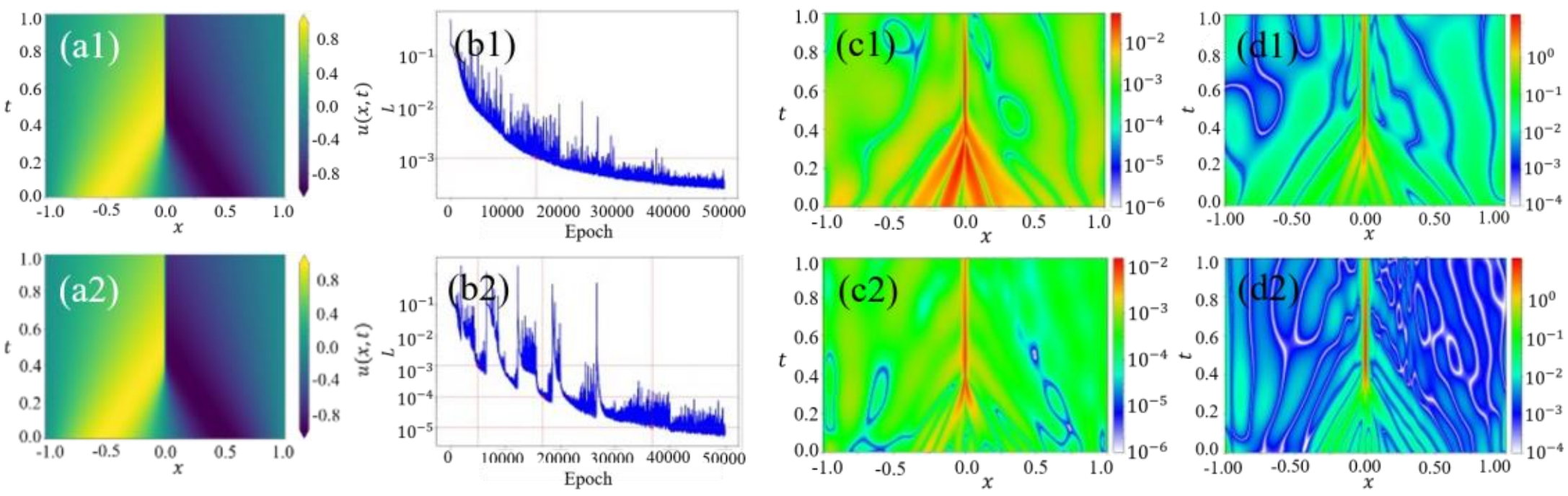

**Fig. 7** Numerical simulation results of Burgers equation (7): reference solution, numerical predictions of baseline PINNs and 3GAS-PINNs under $\eta = 0.7,\ A = 1$. (a1) and (a2) are the predicted solutions of PINNs and 3GAS-PINNs ($\eta = 0.7,\ A = 1$), respectively. (b1) and (b2) are the loss functions of PINNs and 3GAS-PINNs ($\eta = 0.7,\ A = 1$), respectively. (c1), (c2) $e_u$ by PINNs and 3GAS-PINNs, respectively. (d1), (d2) $e_{\nabla u}$ by PINNs and 3GAS-PINNs, respectively.

### 3.4 Korteweg-de Vries equation

The Korteweg-de Vries (KdV) equation is another typical model describing the evolution of waves in dispersive media[27]. It has soliton solutions which are also intermittent structures. Therefore, in this investigation, KdV equation is also simulated to test the performance of 3GAS-PINNs. The normalized KdV equation is [27]

$$u_t + 6uu_x + u_{xxx} = 0, x \in [-1,1], t \in [0,1] \tag{18}$$

where $u(x,t)$ is the wave amplitude in this section.

To evaluate the capability of the proposed method in solving complex nonlinear spatiotemporal dynamics problems, we investigate the interaction of two solitons governed by KdV equation. The analytical solution is constructed via the Hirota bilinear method[38]. The explicit form is given by the second logarithmic derivative[39] of an auxiliary function $f(x,t$:

$$u_{ref}(x,t) = 2\frac{\partial^2}{\partial x^2}\ln f(x,t) \tag{19}$$

The auxiliary function $f(x,t)$ takes the form[40]:

$$f(x,t) = 1 + e^{h_1} + e^{h_2} + A_{12}e^{h_1+h_2} \tag{20}$$

where the variables $h_i$ and the interaction coupling coefficient $A_{12}$ are defined as[41]

$$h_i = \sqrt{c_i}\left(x - c_i t - x_{i,0}\right), \qquad i = 1,2 \tag{21}$$

$$A_{12} = \left(\frac{\sqrt{c_1} - \sqrt{c_2}}{\sqrt{c_1} + \sqrt{c_2}}\right)^2 \tag{22}$$

Here, $c_i$ represents the phase speed of the $i$-th soliton, and $x_{i,0}$ denotes the initial position parameter. To ensure a complete collision event within the computational domain, we set $c_1 = 9.0$ (fast soliton) and $c_2 = 4.0$ (slow soliton), with initial offsets $x_{1,0} = -0.9$ and $x_{2,0} = -0.4$, respectively[42]. A total of 5000 sampling points is employed in the numerical simulations.

The results are shown in Fig. 8. The reference two-soliton solution of Eq. (18) is plotted in Fig. 11 according to Eqs. (19-22), while $|u_x|$ is shown in Fig. 9(a). To explicitly capture these structures, reserving their high-gradient feature, more sampling points are required in the intermittent regime. Fig. 8(b-f) show how $\eta$ controls the distribution of sampling points by 3GAS-PINNs. It can be clearly observed, when $\eta \geq 0.5$, the sampling points concentrate towards the intermittent regime. This is consistent to our expectation.

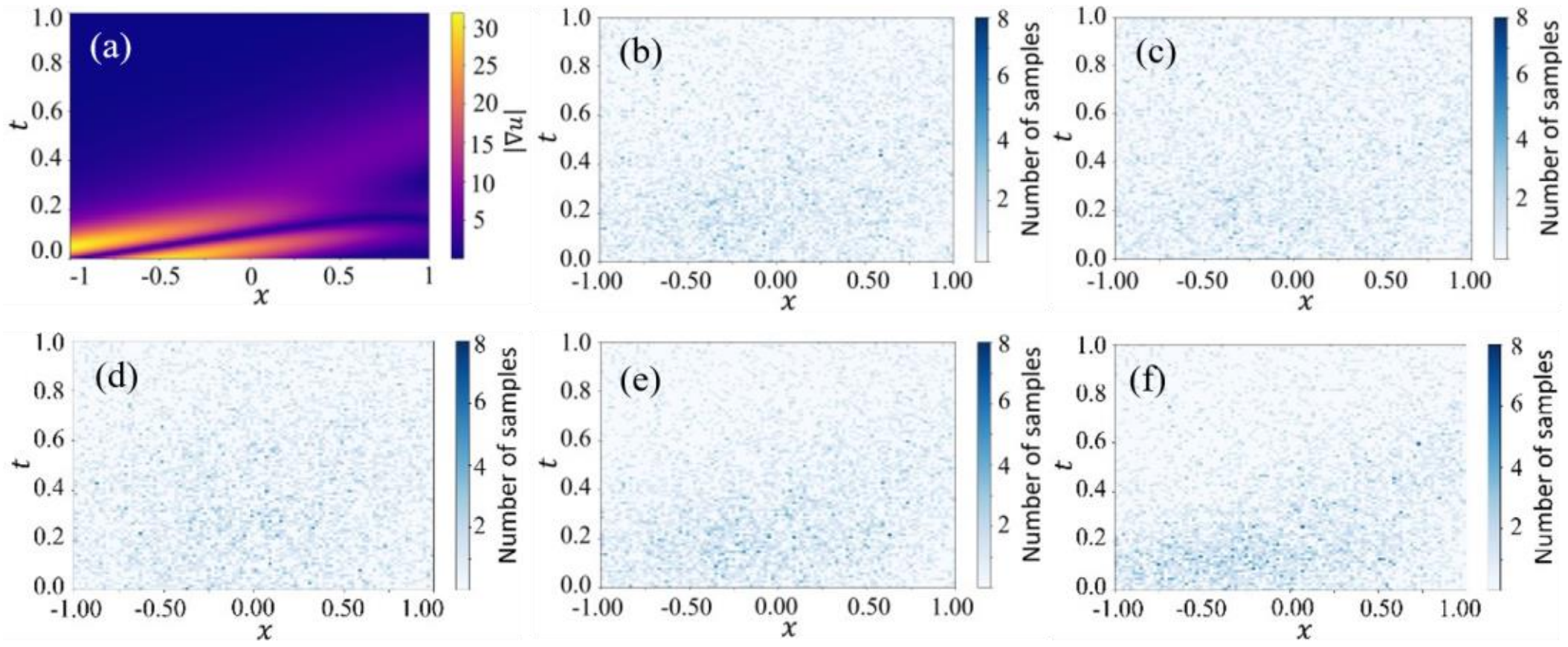

**Fig. 8.** (a) Distribution of gradient magnitude of the reference two-soliton solution of the KdV equation (18). (b) to (f) correspond to the sampling point distributions of the final resampling for the numerical simulation by 3GAS-PINNs at $t = 0.5$, $A = 1$ of the KdV equation (16), with the adaptive update ratio $\eta$ set to 0.1, 0.3, 0.5, 0.7, and 0.9, respectively.

The influence of $\eta$ on simulation accuracy has been summarized in Table 4. When $\eta = 0.7$, $\varepsilon_{L_2}$ and $\varepsilon_{grad,L_2}$ are $2.26 \times 10^{-3}$ and $6.34 \times 10^{-3}$ respectively. They are only 37% and 41% of those in the PINNs ($6.11 \times 10^{-3}$ and $1.55 \times 10^{-2}$). This is equivalent to a factor of $2\sim3$ improvement on simulation accuracy. Notably, excessive adaptation ($\eta = 0.9$) slightly degrades simulation accuracy due to over-sampling in smooth regions. The results support that the 3GAS-PINNs is effective in solving complex soliton interactions governed by the KdV equation.

**Table 4.** Results of 3GAS-PINNs for simulating Equation (18) when $A = 1$ with different values of $\eta$

| $\eta$ | $L(\theta)$ $(10^{-3})$ | $L(\theta)$ $(10^{-4})$ | $L(\theta)$ (Finally) | $\varepsilon_{L_2}$ | $\varepsilon_{grad,L_2}$ |
|---|---|---|---|---|---|
| 0.1 | 9343 | 20915 | 1.41 E-05 | 6.51 E-03 | 1.62 E-02 |
| 0.3 | 10037 | 22088 | 1.49 E-05 | 7.55 E-02 | 1.72 E-02 |
| 0.5 | 10052 | 22104 | 1.86 E-05 | 5.98 E-03 | 1.46 E-02 |
| 0.7 | 11970 | 31584 | 4.12 E-05 | 2.26 E-03 | 6.34 E-03 |
| 0.9 | 9343 | 20915 | 1.41 E-05 | 6.50 E-03 | 1.62 E-02 |
| PINNs | 8780 | 20621 | 1.75 E-05 | 6.11 E-03 | 1.55 E-02 |

We further explore the influence of scaling constant $A$ of the adaptive density function in Eq. (11) on simulation accuracy. As shown in Fig. 9 (a) and (b), the sampling density functions $\rho$ are highly consistent to the distribution of $|u_x|$ in Fig. 8 (a). When $A$ keeps increasing, $\rho$ becomes smeared. The intermittent structure becomes invisible when $A \geq 0.5$. The sampling point distribution also reflects the same tendency of variation.

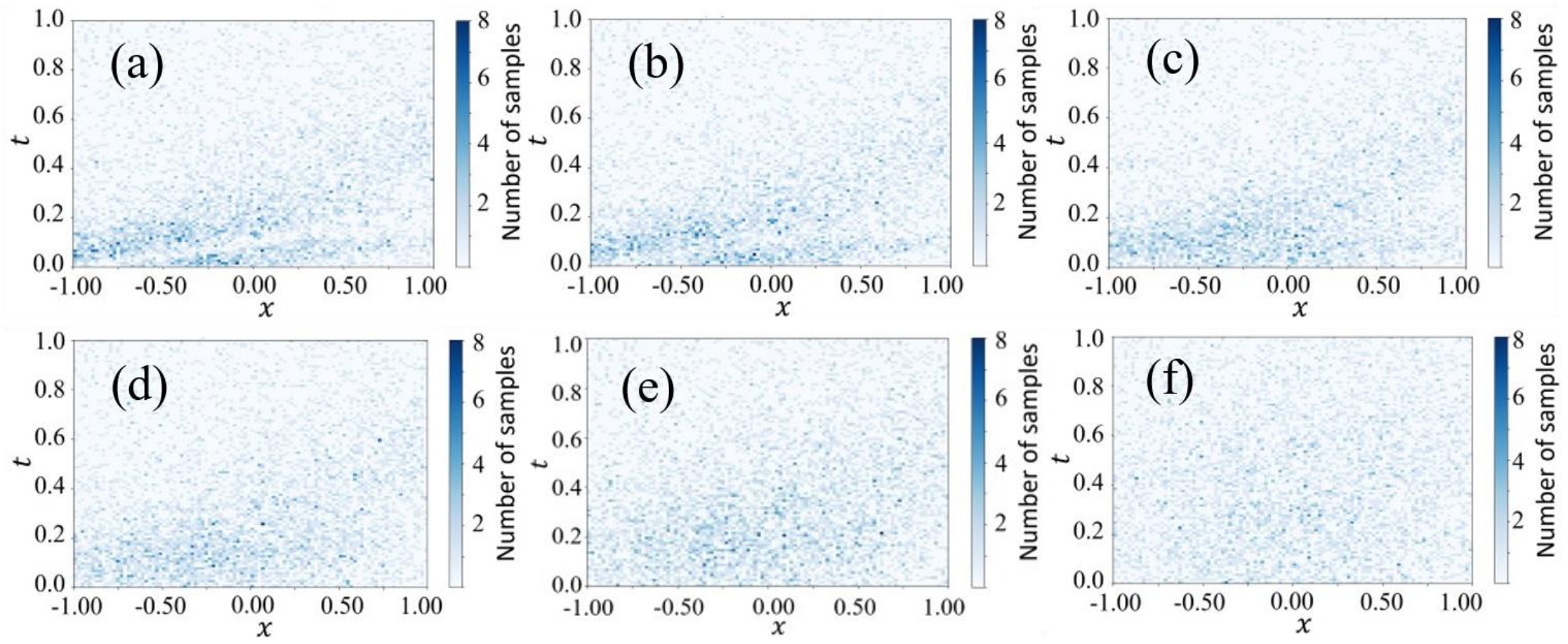

**Fig. 9** Sampling point distribution corresponding to the numerical simulation of the KdV equation (18) by 3GAS-PINNs at $t = 0.5$, with $\eta = 0.7$. (a)–(f) correspond to the final sampling point distributions, with the scaling constant $A$ set to 0.05, 0.1, 0.25, 0.5, 1, and 2, respectively.

On one hand, similar as in the simulation of Burgers equation, the optimal $A$ with the smallest $\varepsilon_{L_2}$ and $\varepsilon_{grad,L_2}$ does not locate at a small value, e.g. 0.05 and 0.1, but around 0.95 and 0.90 (Table 5). The corresponding $\varepsilon_{L_2}$ and $\varepsilon_{grad,L_2}$ are $1.83 \times 10^{-3}$ and $6.07 \times 10^{-3}$ respectively. In contrast, for instance $A = 0.05$, sampling points are concentrated near the two solitons in the intermittent regime, causing a sparse distribution in the smooth regime and low accuracy up there. The $\varepsilon_{L_2}$ and $\varepsilon_{grad,L_2}$ are 3.0 and 2.6 times larger than that of the optimal values.

**Table 5.** Results of the 3GAS-PINNs for simulating Equation (18) when $\eta = 0.7$ with different values of $A$

| $A$ | $L(\theta)$ $(10^{-3})$ | $L(\theta)$ $(10^{-4})$ | $L(\theta)$ (Finally) | $\varepsilon_{L_2}$ | $\varepsilon_{grad,L_2}$ |
|---|---|---|---|---|---|
| 0.05 | 13482 | 34399 | 5.78 E-05 | 5.41 E-03 | 1.59 E-02 |
| 0.10 | 16029 | 36982 | 1.12 E-04 | 7.55 E-02 | 2.06 E-02 |
| 0.25 | 11536 | 30415 | 7.21 E-05 | 5.89 E-03 | 1.58 E-02 |
| 0.50 | 13772 | 37494 | 1.31 E-04 | 5.59 E-03 | 1.23 E-02 |
| 0.90 | 12187 | 27690 | 5.53 E-05 | 6.97 E-03 | 6.07 E-03 |
| 0.95 | 10853 | 23087 | 6.39 E-05 | 1.83 E-03 | 7.71 E-03 |
| 1 | 11970 | 31584 | 4.13 E-05 | 2.27 E-03 | 6.34 E-03 |
| 1.05 | 12037 | 22053 | 4.35 E-05 | 5.58 E-03 | 1.02 E-02 |
| 1.10 | 10909 | 21390 | 5.35 E-05 | 5.72 E-03 | 7.47 E-03 |
| 2 | 10065 | 22324 | 3.65 E-05 | 6.85 E-03 | 1.72 E-02 |
| PINNs | 8780 | 20621 | 1.75 E-05 | 6.11 E-03 | 1.55 E-02 |

On the other hand, unlike the simulations on Burgers equation, 3GAS-PINNs show a relatively smaller accuracy at small number of sampling points, as shown in Fig. 10. Both $\varepsilon_{L_2}$ (Fig. 10 (a)) and $\varepsilon_{grad,L_2}$ (Fig. 10 (b)) are not smaller in the simulation by 3GAS-PINNs, until the number of sampling points is over 2000. Beyond which, the performance of the 3GAS-PINNs becomes surpass the uniform sampling method in PINNs.

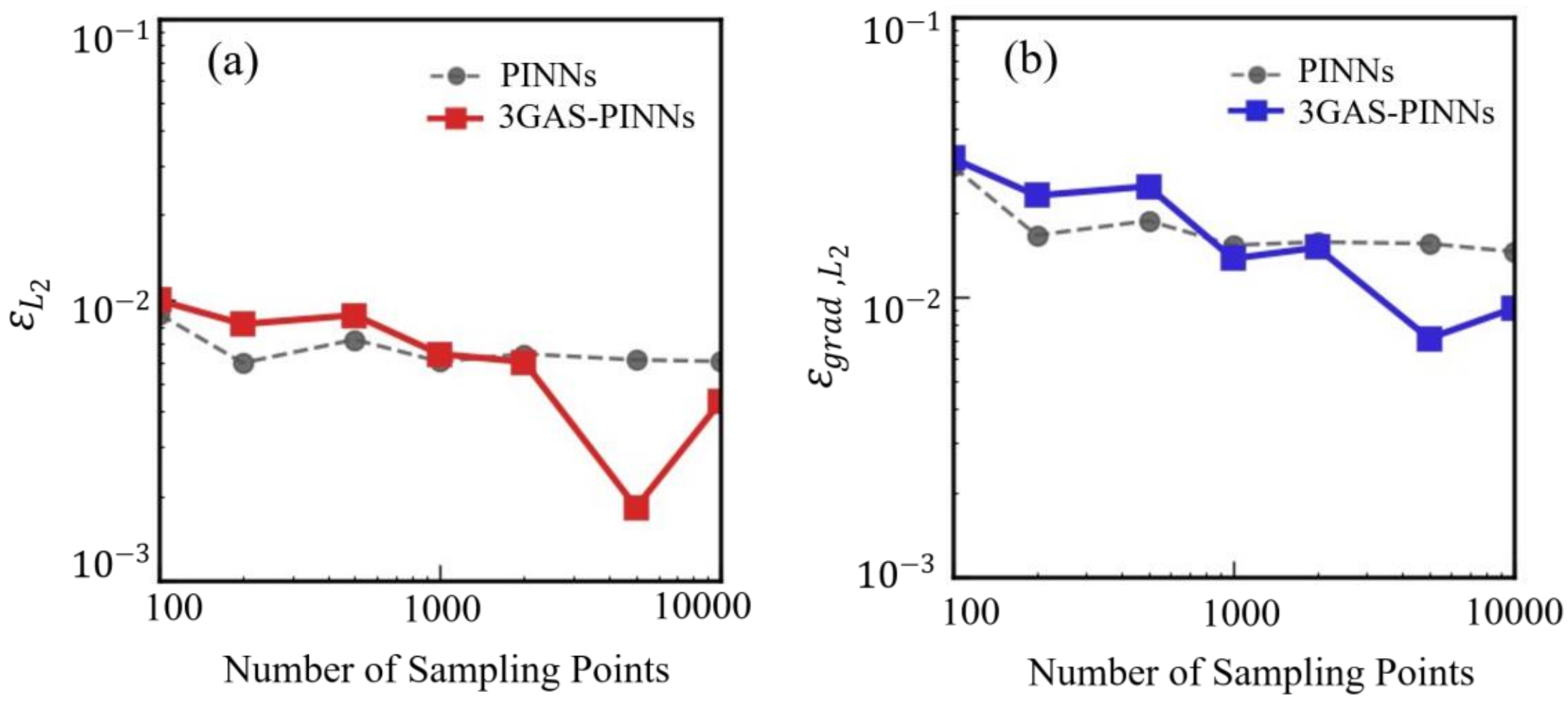

**Fig. 10** Influence of training point number on the errors of solving the KdV equation (18) at $t = 0.5, A = 0.95$ and $\eta = 0.7$ by 3GAS-PINNs and PINNs. (a) $\varepsilon_{L_2}$. (b) $\varepsilon_{grad,L_2}$.

We also conduct a comprehensive performance comparison between the conventional PINNs and the proposed 3GAS-PINNs with optimized hyperparameters ($\eta = 0.7, A = 0.95$) for the KdV equation (18). Superficially, the predicted solutions by PINNs (Fig. 11 (a1)) and 3GAS-PINNs (Fig. 11 (a2)) are consistent to the reference solution. However, the loss functions demonstrate significant difference.

As plotted in Fig. 11 (b1) and (b2), the final training loss of PINNs converges to $1.75 \times 10^{-5}$, whereas 3GAS-PINNs achieves a terminal loss of $6.39 \times 10^{-5}$. In terms of spatial absolute error presented in Fig. 11 (c1) and (c2), PINNs generates widespread error distribution around soliton structures, whereas 3GAS-PINNs dramatically suppress local errors by adaptively enriching collocation points near high-gradient soliton regions via gradient-guided Gaussian Adaptive sampling. Quantitatively, the $\varepsilon_{L_2}$ decreases from $6.11 \times 10^{-3}$ (PINNs) to $1.83 \times 10^{-3}$ (3GAS-PINNs), corresponding to an accuracy improvement by 3 folds. A consistent improvement can also be observed from the absolute gradient error distributions in Fig.11 (d1) and (d2). $\varepsilon_{grad,L_2}$ decreases from $1.55 \times 10^{-2}$ (PINNs) to $7.71 \times 10^{-3}$ (3GAS-PINNs), corresponding to a precision improvement of approximately 2 folds. These quantitative and visual evidences collectively verify that the 3GAS-PINNs can efficiently allocate limited computational resources toward intermittent shock structures and substantially boost the overall numerical precision under identical collocation point budgets.

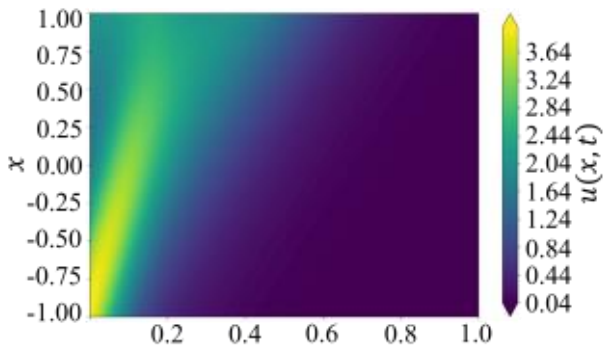

Reference solution

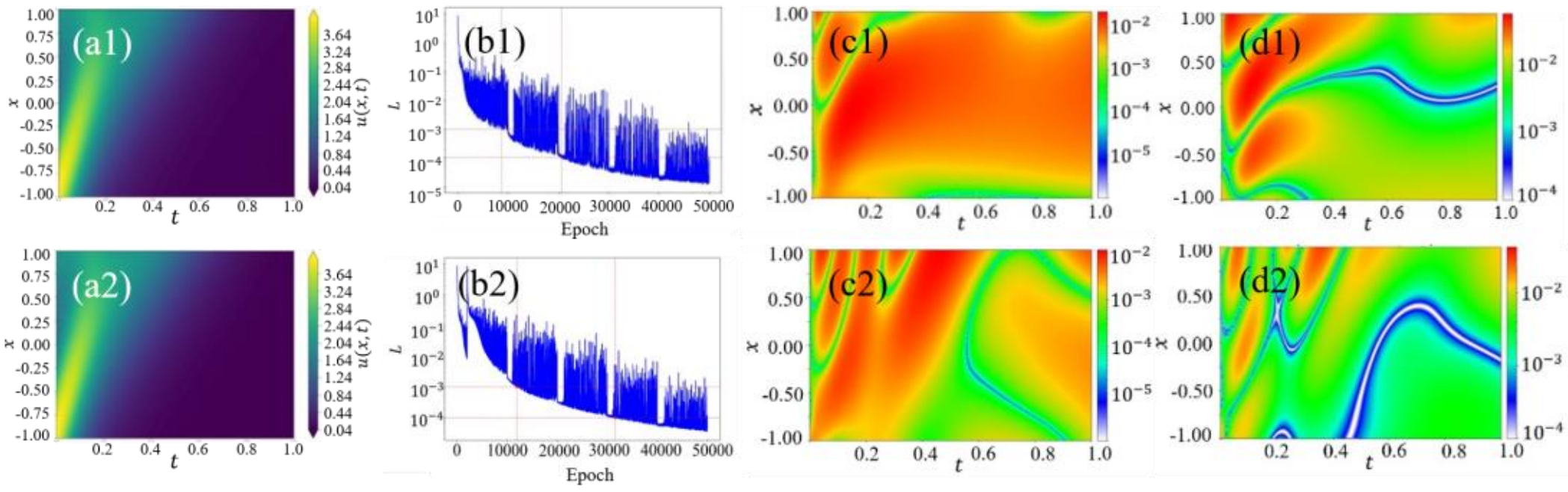

**Fig. 11** Numerical simulation results of KdV equation (18): reference solution, numerical predictions of baseline PINNs and 3GAS-PINNs under $\eta = 0.7,\ A = 0.95$. (a1) and (a2) are the predicted solutions of PINNs and 3GAS-PINNs ($\eta = 0.7,\ A = 0.95$), respectively. (b1) and (b2) are the loss functions of PINNs and 3GAS-PINNs ($\eta = 0.7,\ A = 0.95$), respectively. (c1), (c2) $e_u$ by PINNs and 3GAS-PINNs, respectively. (d1), (d2) $e_{\nabla u}$ by PINNs and 3GAS-PINNs, respectively.

### 3.5 Nonlinear Schrödinger equation

The nonlinear Schrödinger equation (NLSE) is a classical model describing the evolution of wave packets in nonlinear physical systems, e.g. nonlinear fiber optics[27], Bose-Einstein condensation[28] and deepwater waves[29]. Unlike the Burgers and KdV equations which are in real space, nonlinear Schrödinger equation describe wave packets in complex space. The dimensionless form of nonlinear Schrödinger equation is normally given as[43]

$$iu_t + \frac{1}{2}u_{xx} + |u|^2 u = 0 \tag{23}$$

where, $u(x,t)$ denotes the envelope of the complex optical field, $x$ is the longitudinal propagation distance along the fiber, and $t$ represents the retarded time in the reference frame moving with the group velocity. For a continuous-wave background with an amplitude $\Psi_0$, the reference analytical solution of the Peregrine rogue wave is given as [28]

$$u_{ref}(x,t) = u_0 \left| 1 - \frac{4(1 + 2ia^2 t)}{1 + 4a^2x^2 + 4a^4t^2} \right| e^{ia^2 t} \tag{24}$$

where $u_{ref}(x,t)$ denotes the complex-valued analytical solution of the Peregrine rogue wave, $u_0$ represents the amplitude of the continuous-wave background field, $a$ is the characteristic parameter that governs the spatial-temporal distribution and gradient intensity of the rogue wave, $i$ stands for the imaginary unit with $i^2 = -1$, and $x$ and $t$ refer to the spatial coordinate and retarded time, respectively. In this work, $u_0 = 1$, $a = 1$, and the computational domain is $x \in [-5.0, 5.0]$, $t \in [-1.0, 1.0]$.

This solution exhibits an extremely large local gradient at the spatiotemporal central point $(x = 0, t =$

0), posing stringent challenges to the capture accuracy and stability of numerical simulation methods. A total of 5000 sampling points are employed for numerical simulations.

The sampling point distribution are shown in Fig. 12. The reference rogue wave solution of Eq. (23) is plotted in Fig. 14 (a) according to Eq. (24). The rogue wave structure locates at spatiotemporal origin show a drastic variation of $|u_x|$. To explicitly capture these structures, reserving their high-gradient feature, more sampling points are required in the intermittent regime. Fig. 12(b-f) show how $\eta$ controls the distribution of sampling points by 3GAS-PINNs. It can be clearly observed, when $\eta \geq 0.5$, the 3GAS-PINNs concentrates sampling points precisely towards the intermittent regime. This is consistent to our expectation on 3GAS-PINNs.

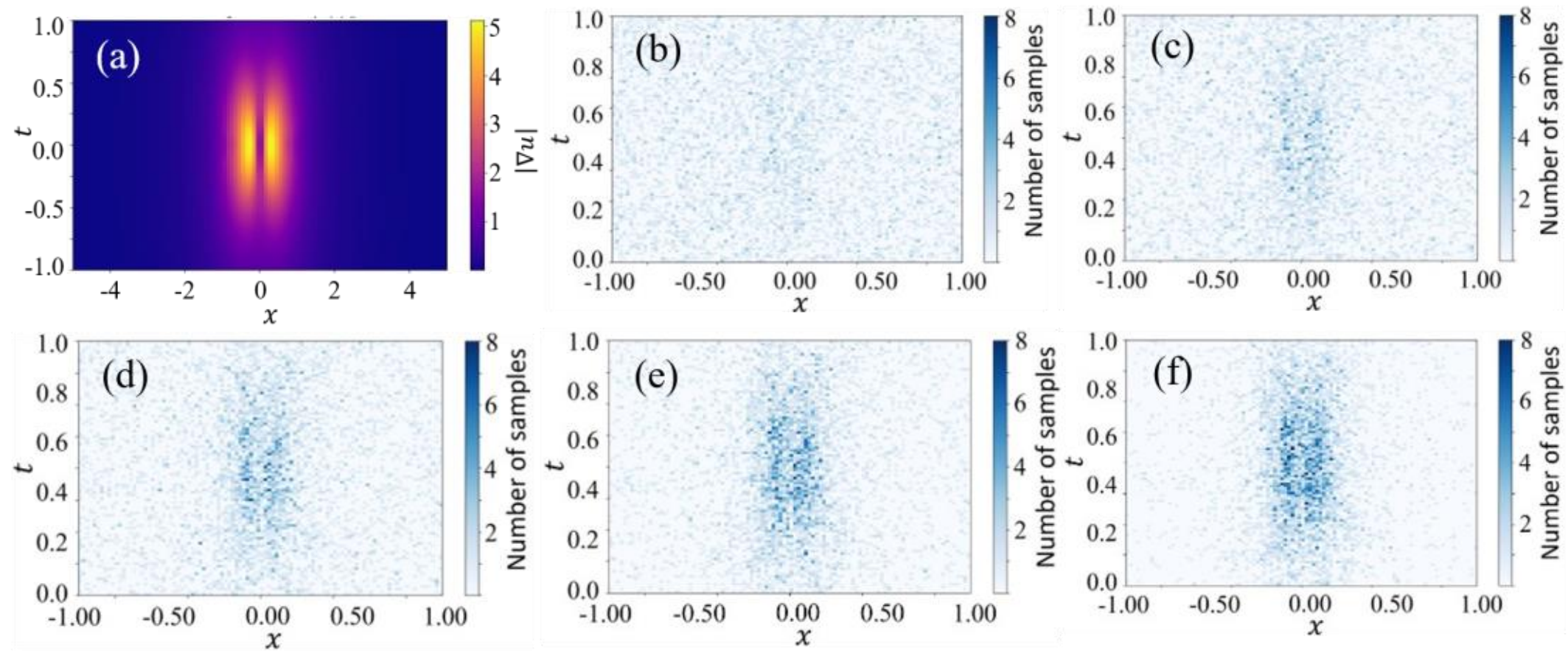

**Fig. 12** Sampling point distribution corresponding to $|u_x|$ (a) Distribution of $|u_x|$ of the reference solution at $t = 0.5$ of the NLSE. (b) to (f) correspond to the sampling point distributions of the final resampling for the numerical simulation by 3GAS-PINNs at $t = 0.5$, $A = 0.3$ of the NLSE, with the adaptive update ratio $\eta$ set to 0.1, 0.3, 0.5, 0.7, and 0.9, respectively.

The influence of $\eta$ on simulation accuracy has been summarized in Table 6. It can be seen the change of $\eta$ will not provide a significant promotion on simulation accuracy. The optimal $\eta = 0.3$, where $\varepsilon_{L_2}$ and $\varepsilon_{grad,L_2}$ are $1.46 \times 10^{-3}$ and $7.18 \times 10^{-3}$ respectively, which are on the equivalent level of those by PINNs ($1.46 \times 10^{-3}$ and $7.23 \times 10^{-3}$). Excessive adaptation ($\eta \geq 0.5$) slightly degrades simulation accuracy due to over-sampling in smooth regions.

**Table 6.** Results of 3GAS-PINNs for simulating Equation (23) when $A = 1$ with different values of $\eta$

| $\eta$ | $L(\theta)$ $(10^{-3})$ | $L(\theta)$ $(10^{-4})$ | $L(\theta)$ (Finally) | $\varepsilon_{L_2}$ | $\varepsilon_{grad,L_2}$ |
|---|---|---|---|---|---|
| 0.1 | 6781 | 19278 | 2.28 E-05 | 1.81 E-03 | 7.75 E-03 |
| 0.3 | 7643 | 20187 | 2.24 E-05 | **1.46 E-03** | **7.18 E-03** |
| 0.5 | 8620 | 21019 | 3.08 E-05 | 1.54 E-03 | 7.45 E-03 |
| 0.7 | 9114 | 21208 | 2.35 E-05 | 1.55 E-03 | 7.63 E-03 |
| 0.9 | 10068 | 22156 | 1.90 E-05 | 2.06 E-03 | 7.55 E-03 |
| PINNs | 8619 | 18747 | 1.60 E-05 | 1.46 E-03 | 7.23 E-03 |

Further exploration on the influence of scaling constant $A$ on simulation accuracy is summarized in Table 7. In this section, the optimal $A = 0.1$, which provides the lowest $\varepsilon_{L_2}$ and $\varepsilon_{grad,L_2}$ at $1.31 \times 10^{-3}$ and $7.17 \times 10^{-3}$ respectively. Unfortunately, the promotion is also limited. Unlike the optimal $A$ in Burgers and KdV equations, the optimal $A$ in the simulation of NLSE is much lower. The sampling point distribution also shows a high consistency with $|u_x|$ (Fig. 12(a)).

**Table 7.** Results of 3GAS-PINNs for simulating Equation (23) when $\eta = 0.3$ with different values of $A$

| $A$ | $L(\theta)$ $(10^{-3})$ | $L(\theta)$ $(10^{-4})$ | $L(\theta)$ (Finally) | $\varepsilon_{L_2}$ | $\varepsilon_{grad,L_2}$ |
|---|---|---|---|---|---|
| 0.05 | 7809 | 20036 | 1.69 E -05 | 1.36 E -03 | 7.13 E -02 |
| 0.1 | 7472 | 20175 | 2.36 E -05 | **1.31 E -03** | **7.17 E -03** |
| 0.25 | 7742 | 20087 | 4.14 E -05 | 1.70 E -03 | 7.19 E -03 |
| 0.5 | 7643 | 20187 | 2.42 E -05 | 1.46 E -03 | 7.18 E -02 |
| 1 | 7711 | 20077 | 2.04 E -05 | 1.44 E -03 | 7.19 E -02 |
| 2 | 7357 | 20055 | 1.95 E -05 | 1.46 E -03 | 7.17 E -02 |
| PINNs | 8619 | 18747 | 1.60 E -05 | 1.46 E -03 | 7.23 E -03 |

Fig. 13 demonstrate that the 3GAS-PINNs can reduce the $\varepsilon_{L_2}$ and $\varepsilon_{grad,L_2}$ by up to one order of magnitude relative to PINNs, particularly at a small number of sampling points. For instance, when only 100 sampling points are employed, $\varepsilon_{L_2}$ and $\varepsilon_{grad,L_2}$ by 3GAS-PINNs are 14 and 11 folds higher than that of PINNs. The performance of the 3GAS-PINNs over standard PINNs becomes increasingly prominent as the number of sampling points decreases. More importantly, for the numerical simulation of the nonlinear Schrödinger equation, the 3GAS-PINNs maintains consistently lower $\varepsilon_{L_2}$ and $\varepsilon_{grad,L_2}$ across all sampling budgets. This confirms the remarkable efficiency of the adaptive sampling strategy when computational resources are constrained.

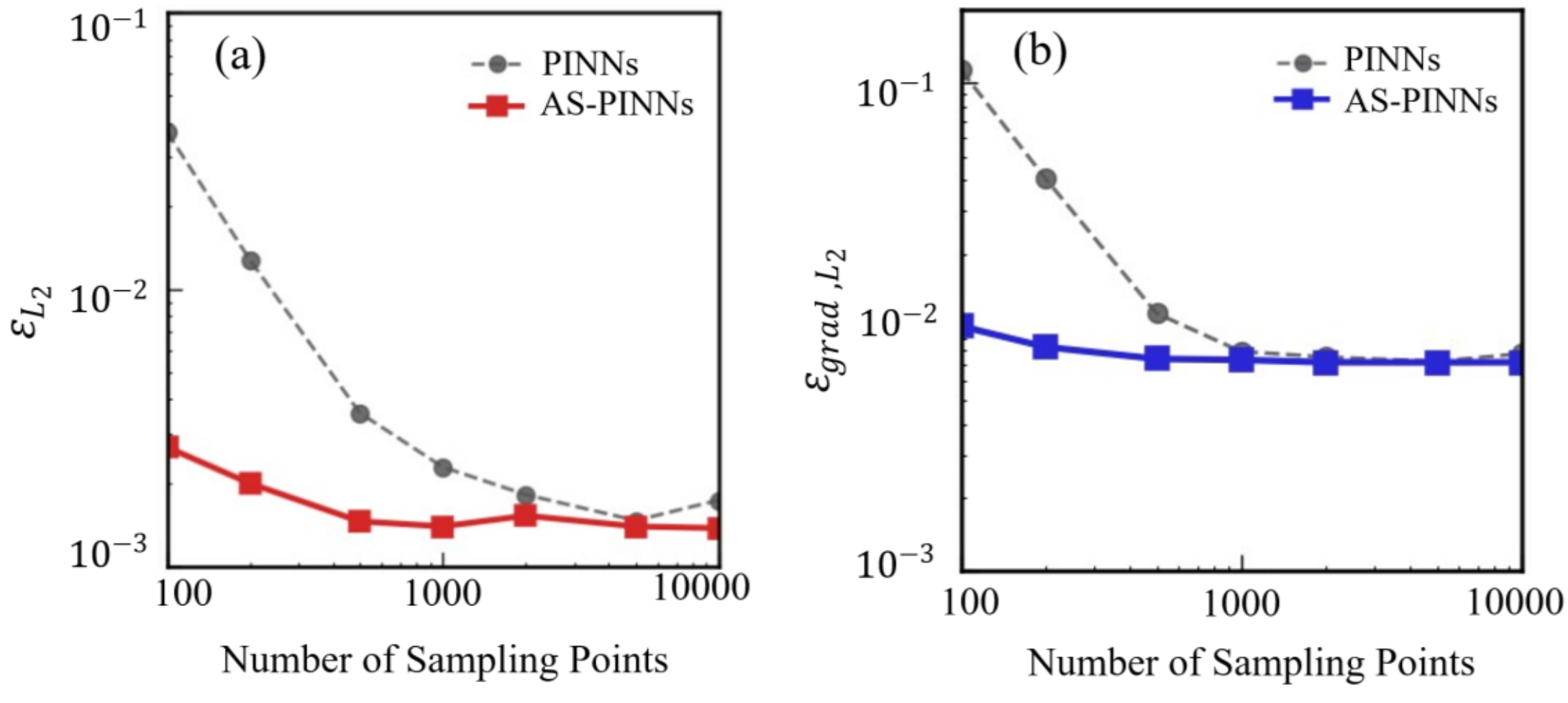

**Fig. 13** Influence of training point number on the errors of solving NLSE by 3GAS-PINNs and PINNs. (a) $\varepsilon_{L_2}$. (b) $\varepsilon_{grad,L_2}$.

The comparisons among the reference solution and the prediction solutions of conventional PINNs and 3GAS-PINNs with optimized hyperparameters ($\eta = 0.3, A = 0.01$) for solving the NLSE are shown in Fig. 14. The predicted solutions by PINNs (Fig. 14 (a1)) and 3GAS-PINNs (Fig. 14 (a2)) are consistent to the theoretical solution as well. As illustrated in the loss evolution plots (Fig. 14 (b1) and (b2)) the final training loss of PINNs converges to $1.60 \times 10^{-5}$(Fig. 14 (b1)), while 3GAS-PINNs achieves a terminal loss of $2.36 \times 10^{-5}$ (Fig. 14 (b2)). In terms of absolute error $\Delta u$ presented in Fig.14. (c1) and Fig.14 (c2), PINNs generates widespread error distribution around soliton structures, whereas 3GAS-PINNs dramatically suppress local errors by adaptively enriching collocation points near high-gradient soliton regions via the adaptive sampling strategy. A consistent improvement can also be observed from the absolute gradient error distributions in Fig.14 (d1) and Fig.14 (d2). Quantitatively, the $\varepsilon_{grad,L_2}$ decreases from $7.23 \times 10^{-3}$ (PINNs) to $7.17 \times 10^{-3}$ (3GAS-PINNs). These quantitative and visual evidences again verify that the adaptive sampling scheme embedded in 3GAS-PINNs can efficiently allocate limited computational resources toward intermittent shock structures and substantially boost the overall numerical precision under identical collocation point budgets.

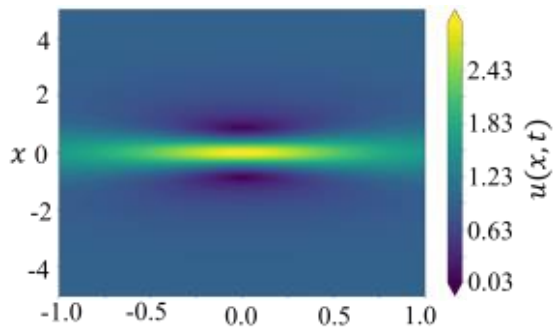

Reference solution

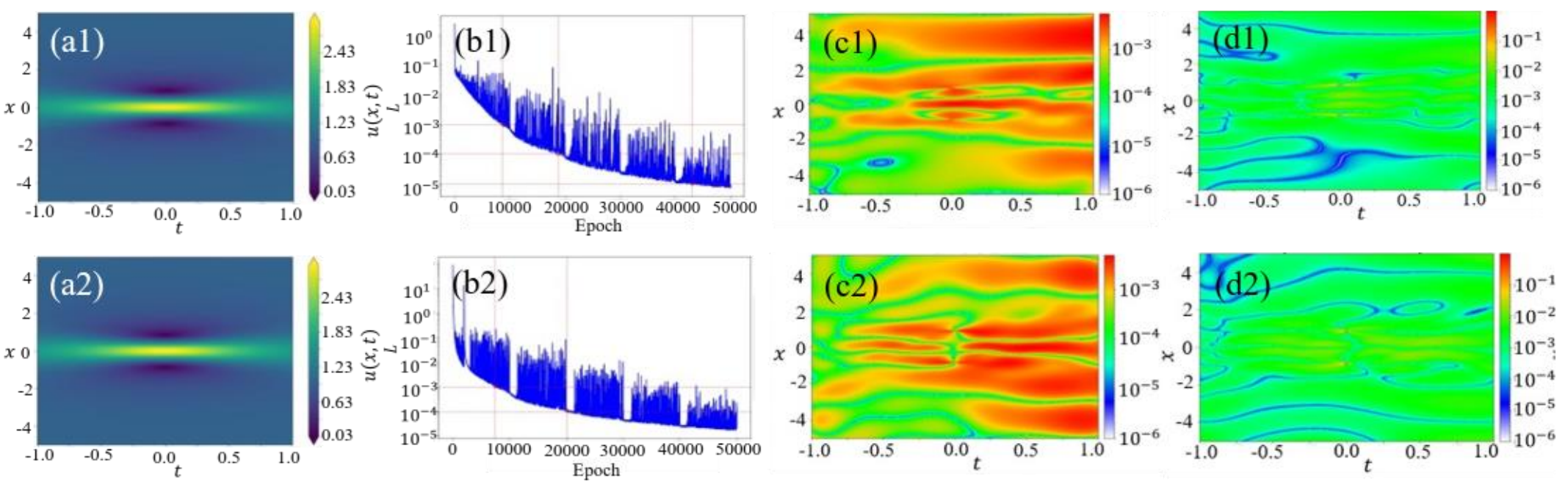

**Fig. 14.** Numerical simulation results of NLS equation (23): reference solution, numerical predictions of baseline PINNs and 3GAS-PINNs under $\eta = 0.3$, $A = 0.1$. (a1) and (a2) are the predicted solutions of PINNs and 3GAS-PINNs ($\eta = 0.7$, $A = 1$), respectively. (b1) and (b2) are the loss functions of PINNs and 3GAS-PINNs ($\eta = 0.7$, $A = 1$), respectively. (c1), (c2) $e_u$ by PINNs and 3GAS-PINNs, respectively. (d1), (d2) $e_{\nabla u}$ by PINNs and 3GAS-PINNs, respectively.

## 4. Discussion

The accuracy gains of 3GAS-PINNs across three representative nonlinear PDEs derive from a closed-loop physical feedback workflow that alleviates gradient lag existing in uniformly sampled baseline PINNs. Conventional uniform collocation distributes identical sampling budgets over smooth domains and sharp intermittent regions. Low-gradient background areas dominate cumulative training loss during optimization, driving the network to recover low-frequency solution components, whereas collocation points are sparsely distributed around shock, soliton and rogue wave fronts, yielding smeared transition structures and measurable deviations relative to reference solutions. The gradient-guided Gaussian sampling process calculates real-time spatial gradient magnitudes via automatic differentiation to characterize local solution intermittency, and smooths discrete gradient spikes into a continuous regional field using Gaussian kernel convolution. It further produces physically consistent sampling probability density, and hybrid sampling integrates adaptive refinement for high-gradient regions with full-domain uniform collocation to sustain global PDE constraints. Dynamic redistribution of collocation points rebalances the local intermittency, accelerates gradient propagation across steep interfaces, and enables simultaneous reconstruction of large-scale background flow and microscale intermittent structures without breaking full-domain physical consistency.

Parametric investigations show distinct optimal values of the Gaussian scaling coefficient $A$ across the Burgers, KdV and nonlinear Schrödinger equations. This discrepancy arises from the inherent multiscale balance of each nonlinear system, instead of random training fluctuations. The kernel width Eq. (11) controls the spatial scope of intensified sampling near high-gradient zones. For Burgers flows with shock fronts, $A = 1$ yields balanced collocation coverage over discontinuities and surrounding smooth fields. Smaller $A$ values concentrate collocation points excessively around gradient peaks and leave smooth domains undersampled, while overly large $A$ blurs gradient information and weakens targeted refinement. KdV equations produce moving solitons, and these solitons form broad collision zones with steep gradients. We select $A = 0.95$ as the optimal scaling factor for the Gaussian kernel. This value widens the kernel's effective coverage to fully cover the continuous transition band of soliton collisions. The sampling layout then changes smoothly alongside gradient shifts across the whole collision zone. This arrangement stops collocation points from clustering only at single gradient peaks. Instead, it evenly refines all high-gradient areas within the soliton collision region. In contrast, the NLSE supports Peregrine rogue waves with tightly confined amplitude singularities and high-frequency envelopes, which has a small optimal $A = 0.1$ to shrink kernel range and centralize collocation resources on the gradient spike at the spatiotemporal origin. Larger $A$ disperses sampling density onto peripheral smooth regions and lowers the reconstruction quality of rogue wave amplitude and gradients. Overall, the optimal $A$ is determined by the characteristic spatial scale of each problem's intermittent structures and the effective spatial extend over which gradients drive obvious variations in the solution field.

In recent years, several adaptive sampling methods have been developed from different aspects. For instance, residual-based sampling methods construct probability distributions using PDE residual magnitudes as indirect evaluation metrics[20, 21]. Residual-based sampling methods build sampling probabilities using PDE residual values. However, numerical noise heavily distorts these residuals at the start of training, creating false high residual readings in smooth, low-gradient flow. As a result, the algorithm places too many collocation points in these smooth areas, while intermittent structures such as shock waves and solitons cannot get sufficient sampling refinement. It also suffers temporal delay when solving moving wave structures, since supplementary collocation points are only arranged after obvious residuals accumulate. In contrast, our method directly adopts the real-time gradient field of predicted solutions as the direct indicator of local structure, avoiding indirect error feedback lag. The Gaussian convolution regularizes discrete gradient spikes into continuous weight fields, which suppresses invalid point aggregation in flat regions. Besides, our gradient information captures solution variations synchronously and completes point redistribution in each resampling cycle without delayed response to traveling intermittent features. Relative to the metaheuristic adaptive approaches, such as FAMAW-PINNs, which require exhaustive full spatio-temporal domain traversal during each resampling iteration with redundant computational costs[19], our method achieves targeted collocation based on local structures without global domain searching, delivering higher sampling efficiency.

Under equal collocation budgets, the relative $L_2$ error of Burgers shocks drops to 14.6% of conventional PINNs. For KdV colliding solitons, the relative solution and gradient $L_2$ errors are reduced by 3 folds and 2 folds, respectively. Even with sparse collocation resources, the solution error $\varepsilon_{L_2}$ and gradient error $\varepsilon_{grad,L_2}$ of 3GAS-PINNs are only 1/14 and 1/11 of those yielded higher prediction accuracy for the extreme gradient singularities of NLSE rogue waves by conventional PINNs.

## 5. Conclusion

In this study, a PINN framework integrating adaptive sampling mechanisms are proposed. The proposed adaptive sampling method successfully captures complex physical features. By using the gradient information of the current solution, collocation points are concentrated in regions with large variations. The adaptive weighting strategy balances the significance of different loss terms. This mechanism mitigates the gradient hysteresis in the PINNs. Numerical experiments on the KdV equation demonstrate that adaptive sampling can improve the predicted solution accuracy by 2–3 times, showing its superiority in traveling wave model. Numerical experiments on the nonlinear Schrödinger equation demonstrate that adaptive sampling strategy achieves a 2 to 14-fold increase in accuracy over traditional PINNs, showing its accurate solutions can be obtained using significantly fewer computational resources.